\documentclass[11pt,a4paper]{article}
\usepackage{etex}  
\usepackage[hyperref]{emnlp2018}
\usepackage{standalone}
\usepackage{times}
\usepackage{latexsym}
\usepackage{url}
\usepackage{amsmath}
\usepackage{amssymb}
\usepackage{amsfonts}
\usepackage{graphicx}
\usepackage[skip=5pt]{caption}
\usepackage{subcaption}
\usepackage{array}
\usepackage{tabu}
\usepackage{makecell}
\usepackage{subcaption}
\usepackage{paralist}
\usepackage{cases}
\usepackage{diagbox}
\usepackage{enumitem}
\usepackage{soul}
\usepackage{multirow}
\usepackage{verbatim}
\usepackage{tabulary}
\usepackage{booktabs}
\usepackage[mathscr]{euscript}
\usepackage{mathtools}
\usepackage{amssymb}
\usepackage{algorithm}
\usepackage{algpseudocode}
\usepackage{stmaryrd}
\usepackage{amsthm}
\usepackage{tikz-dependency}
\usetikzlibrary{automata,decorations.markings,arrows,positioning,matrix,calc,patterns,angles,quotes,calc}
\usepackage{adjustbox}
\usepackage[toc,page]{appendix}
\usepackage{tabularx}
\usepackage{xspace}

\DeclarePairedDelimiter\abs{\lvert}{\rvert}%

\algrenewcommand{\algorithmiccomment}[1]{\leavevmode\hfill$\triangleright$ #1}

\usepackage{color}
\usepackage{bm}
\definecolor{orange}{rgb}{1,0.5,0}
\definecolor{mdgreen}{rgb}{0.05,0.6,0.05}
\definecolor{mdblue}{rgb}{0,0,0.7}
\definecolor{dkblue}{rgb}{0,0,0.5}
\definecolor{dkgray}{rgb}{0.3,0.3,0.3}
\definecolor{slate}{rgb}{0.25,0.25,0.4}
\definecolor{gray}{rgb}{0.5,0.5,0.5}
\definecolor{ltgray}{rgb}{0.7,0.7,0.7}
\definecolor{purple}{rgb}{0.7,0,1.0}
\definecolor{lavender}{rgb}{0.65,0.55,1.0}

\newcommand{\ensuretext}[1]{#1}
\newcommand{\marker}[2]{\ensuremath{^{\textsc{#1}}_{\textsc{#2}}}}

\newcommand{\arkcomment}[3]{\ensuretext{\textcolor{#3}{[#1 #2]}}}
\renewcommand{\arkcomment}[3]{}  

\newcommand{\isubsection}[2]{\subsection{#1}\label{sec:#2}}

\newcommand{\figref}[1]{Figure~\ref{fig:#1}}
\newcommand{\tabref}[1]{Table~\ref{tab:#1}}
\newcommand{\secref}[1]{\S\ref{sec:#1}}
\newcommand{\appref}[1]{Appendix~\ref{sec:#1}}
\newcommand{\nascomment}[1]{\arkcomment{\marker{NA}{S}}{#1}{blue}}
\newcommand{\hao}[1]{\arkcomment{\marker{H}{P}}{#1}{mdgreen}}
\newcommand{\roy}[1]{\arkcomment{\marker{R}{S}}{#1}{purple}}
\newcommand{\sam}[1]{\arkcomment{\marker{S}{T}}{#1}{orange}}
\newcommand{\com}[1]{}
\newcommand{\resolved}[1]{}
\newcommand{\camready}[1]{#1}

\newcommand{\wfsa}[1]{\mathscr{#1}}

\newcommand{\repo}{\url{https://github.com/Noahs-ARK/rational-recurrences}}
\newcommand{\unigram}{\textsc{rrnn}($\wfsa{B}$)\xspace}
\newcommand{\unimp}{\textsc{rrnn}($\wfsa{B}$)$_{\text{m+}}$\xspace}
\newcommand{\bigram}{\textsc{rrnn}($\wfsa{C}$)\xspace}
\newcommand{\both}{\textsc{rrnn}($\wfsa{F}$)\xspace}
\newcommand{\lstm}{LSTM\xspace}

\newcommand{\term}[1]{\textbf{#1}} 
\newcommand{\seq}[1]{\mathbf{#1}}

\newcommand{\K}{\mathbb{K}}

\newcommand{\epstrans}{$\varepsilon$\=/transition\xspace{}}
\newcommand{\interalia}[1]{\citep[\emph{inter alia}]{#1}}

\newcommand{\names}[0]{rational recurrences\xspace}

\newcolumntype{L}[1]{>{\raggedright\let\newline\\\arraybackslash\hspace{0pt}}m{#1}}
\newcolumntype{C}[1]{>{\centering\let\newline\\\arraybackslash\hspace{0pt}}m{#1}}
\newcolumntype{R}[1]{>{\raggedleft\let\newline\\\arraybackslash\hspace{0pt}}m{#1}}

\newtheorem{theorem}{Theorem}

\newtheorem{proposition}[theorem]{Proposition}
\newtheorem{corollary}[theorem]{Corollary}
\theoremstyle{definition}
\newtheorem{definition}[theorem]{Definition}
\newtheorem{example}[theorem]{Example}
\theoremstyle{remark}

\setul{1pt}{.4pt}

\DeclareCaptionFont{tiny}{\tiny}

\usepackage{url}
\usepackage[shortcuts]{extdash}  

\aclfinalcopy 

\newcommand\ourtitle{Rational Recurrences}

\title{\ourtitle}

\author{Hao Peng$^\diamondsuit$ \quad Roy Schwartz$^\diamondsuit$$^\heartsuit$ \quad
  Sam Thomson$^\clubsuit$  \quad
  Noah A. Smith$^\diamondsuit$$^\heartsuit$ \\
  $^\diamondsuit$Paul G. Allen School of Computer Science \& Engineering,
  University of Washington, Seattle, WA, USA \\
  $^\clubsuit$Language Technologies Institute,
  Carnegie Mellon University, Pittsburgh, PA, USA \\
  $^\heartsuit$Allen Institute for Artificial Intelligence, Seattle, WA, USA \\
  {\tt \{hapeng,roysch,nasmith\}@cs.washington.edu,
    sthomson@cs.cmu.edu}}

\date{}

\begin{document}
\setlength{\abovedisplayskip}{7pt}
\setlength{\belowdisplayskip}{7pt}
\maketitle
\begin{abstract}
Despite the tremendous empirical success of neural models in
natural language processing,
many of them lack the strong intuitions that accompany classical machine learning approaches.
Recently, connections have been shown between convolutional neural networks (CNNs) and
weighted finite state automata (WFSAs), leading to new interpretations and insights.
In this work, we show that some \emph{recurrent} neural networks also share this connection to WFSAs.
We characterize this connection formally, defining \term{\names} to be
recurrent hidden state update functions that can be written as
the Forward calculation of a finite set of WFSAs.
We show that several recent neural models use \names.
Our analysis provides a fresh view of these models and 
facilitates devising new neural architectures that draw inspiration from WFSAs.
We present one such model, which performs better than two recent baselines
on language modeling and text classification.
Our results demonstrate that transferring intuitions from classical models
like WFSAs can be an effective approach to designing and understanding neural models.
\end{abstract}

\section{Introduction}
\label{sec:introduction}

Neural models, and in particular  gated variants of recurrent neural networks (RNNs, e.g., \citealp{Hochreiter:1997,Cho:2014}), have become a core building block
for state-of-the-art approaches in NLP \cite{Goldberg:2016}.
While these models empirically outperform classical NLP methods on many tasks~\interalia{zaremba2014recurrent,bahdanau2014neural,dyer2016rnng,peng2017deep}, 
they typically lack the intuition offered by classical models, 
making it hard to understand the roles played by each of their components.
In this work we show that many neural models are more interpretable than previously thought,
by drawing connections to weighted finite state automata (WFSAs).
We study several recently proposed RNN architectures and show 
that one can use WFSAs to characterize their recurrent updates.
We call such models \term{\names}~(\secref{rnn2wfsa}).%
\footnote{
	Where the term \emph{regular} is used with unweighted FSAs (e.g., regular languages,
	regular expressions), \emph{rational} is the weighted analog (e.g., rational
	series, \citealp{sakarovitch_rational_2009}; rational
	kernels,~\citealp{cortes_rational_2004}).
}
Analyzing recurrences in terms of WFSAs provides a new view of existing
models and facilitates the development of new ones.

\begin{figure}
	\centering
	\includegraphics[clip,trim=8cm 9.5cm 8cm 9.5cm, width=.8\columnwidth]{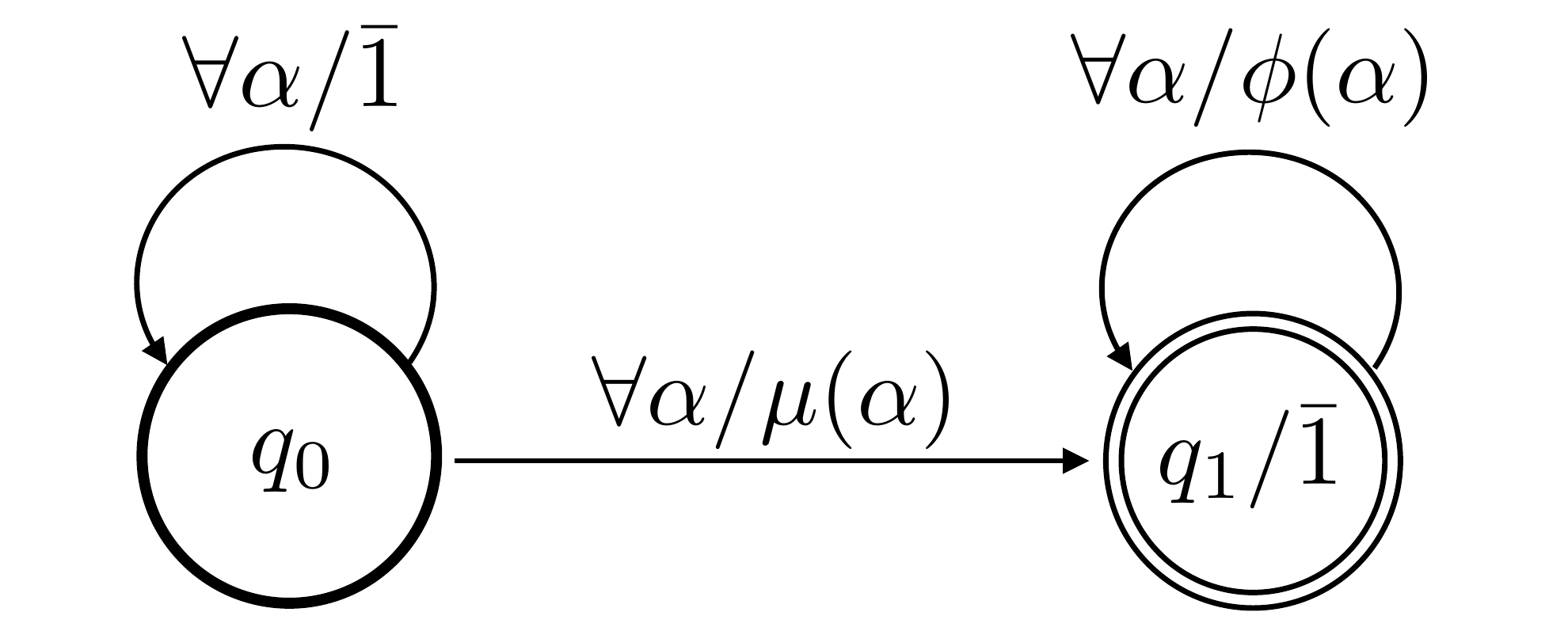}
	\caption{A two-state WFSA $\wfsa{B}$ described in \secref{background}.
		It is closely related to several models studied in this paper (\secref{neural}). 
		Bold circles indicate initial states,
		and double circles final states, which are associated with final weights.
		Arrows represent transitions, labeled by the symbols $\alpha$ they consume,
		and the weights as a function of $\alpha$.
		Arcs not drawn are assumed to have weight $\bar{0}$.
		For brevity,~$\forall\alpha$ means~$\forall\alpha\in\Sigma$, with $\Sigma$ being the alphabet.}
	\label{fig:wfsa:b}
\end{figure}

In recent work,~\citet{Schwartz:2018} introduced SoPa, an RNN 
constructed from WFSAs, and thus rational by our definition.
They also showed that a single-layer max-pooled CNN \cite{Lecun:1998}
can be simulated by a set of simple WFSAs (one per output dimension), and accordingly are also rational.
In this paper we broaden such efforts, and show that \names are in frequent use%
\sam{give the name of the model next to the citation?}%
~\citep{mikolov2014learning,Balduzzi:2016,Lei:2016,Lei:2017a,Lei:2017b,Bradbury:2017,Foerster:2017}.
For instance, we will show in~\secref{relations} that
the WFSA diagrammed in~\figref{wfsa:b} 
has strong connections to several of the models mentioned above.


Based on these observations, we then discuss potential approaches to deriving 
novel neural architectures from  WFSAs (\secref{derive}).
As a case study, we present a new model motivated by 
the interpolation of a two-state WFSA and a three-state one,
capturing (soft) unigram and bigram features, respectively.
Our experiments show that in two tasks---language modeling and text classification---the 
proposed model outperforms recently proposed 
rational models~(\secref{experiments}).
Further extensions might lead to larger gains, and
the rational recurrence view could facilitate easier exploration of such extensions.
To promote such exploration, we publicly release our implementation at \repo.

\section{Background: Weighted Finite State Automata (WFSAs)}
\label{sec:background}
\resolved{\hao{we use \term{term} for some terms, and \textit{term} for others.
do we want to keep them consistent?}}
This section reviews weighted finite-state automata and semirings, which
underly our analyses in \secref{rnn2wfsa}.
WFSAs extend nondeterministic \emph{unweighted} finite-state automata by assigning
weights to transitions, start states, and final states.
Instead of simply accepting or rejecting a string, a WFSA returns a
score for the string,
and this score summarizes the weights along all paths through the WFSA
that consume the string.
In order for this summary score to be efficiently computable,
weights are taken from a \term{semiring}.

\begin{definition}[\citealp{kuich1985semiring}]
A \textbf{semiring} is a set $\K$ along with two associative binary operations
on $\K$, $\oplus$ (addition) and  $\otimes$ (multiplication), and two identity
elements: $\bar{0}$ for addition, and $\bar{1}$ for multiplication.
Semirings also require that addition is commutative, multiplication distributes
over addition, and that multiplication by $\bar{0}$ annihilates
(i.e.,~$\bar{0}\otimes a = a \otimes \bar{0} = \bar{0}$).
\end{definition}
One common semiring is the \term{real} (or plus-times) semiring:~$\left\langle\mathbb{R}, +, \cdot, 0,1\right\rangle$.
The other one used in this work is the \term{max-plus} semiring~$\left\langle\mathbb{R}\cup\{-\infty\}, \max, +, -\infty,0\right\rangle$.
We refer the reader to \citet{kuich1985semiring}
for others.

\begin{definition}
A \term{weighted finite-state automaton (WFSA)} over a semiring $\mathbb{K}$
is a 5-tuple,~$\wfsa{A}=\left\langle \Sigma, \mathcal{Q}, \tau, \lambda, \rho\right\rangle$,\footnote{%
Some authors define $\tau$, $\lambda$, and $\rho$ to be partial functions---applying only a subset of transitions, initial states, and final states respectively.
Our definition is equivalent, giving the weight functions value $\bar{0}$ wherever they were undefined.
}
with:
\begin{compactitem}[$\bullet$]
	\item a finite input alphabet $\Sigma$;
	\item a finite state set $\mathcal{Q}$;
	\item transition weights $\tau:\mathcal{Q}\times\mathcal{Q} \times \left(\Sigma\cup\{\varepsilon\}\right) \rightarrow \mathbb{K}$;
	\item initial weights $\lambda:\mathcal{Q} \rightarrow \mathbb{K}$;
	\item and final weights $\rho:\mathcal{Q} \rightarrow \mathbb{K}$.
\end{compactitem}
\end{definition}
\noindent 
$\varepsilon\notin \Sigma$~marks
special \term{\epstrans s} that may be taken without consuming any input.
$\wfsa{A}$ assigns a score $\wfsa{A}\llbracket \seq{x} \rrbracket$ to a string~$\seq{x} = x_1 \ldots x_n \in \Sigma^\ast$~by summing over the scores
of all possible paths deriving $\seq{x}$.
The score of each individual path is the product of the weights of the transitions it consists of. Formally:
\begin{definition}[path score]
\label{def:path-weight}
Let $\bm{\pi}=\pi_1 \ldots \pi_n$ be a sequence of adjacent \term{transitions} in
$\wfsa{A}$, with each transition
$\pi_i = (q_i, q_{i+1}, z_i) \in \mathcal{Q} \times \mathcal{Q} \times \left(\Sigma\cup\{\varepsilon\}\right)$.
The path $\bm{\pi}$ \term{derives} string $\seq{x} \in \Sigma^*$, which is the substring of $\seq{z}=z_1z_2\dots z_n$ that excludes $\varepsilon$ symbols
(for example, if $\seq{z} = a \varepsilon b c \varepsilon \varepsilon \varepsilon d$, then $\seq{x} = a b c d$).
\resolved{
\roy{I remember having issues with this sentence before submission. I am still not entirely sure I understand what it means. Are you only considering paths that do not include $\epsilon$ paths? if so, fix the definition of $\pi_i$ to exclude them. Are you simply referring to the subset of $x$ captured by $\bm{\pi}$? than it's clear that $\epsilon$s are not part of it.}
\hao{i guess here it means that one can take a $\varepsilon$ transition, but the $\varepsilon$-symbol does not go into the substring we care about.
and there is indeed inconsistency in the notation here: in the above paragraph,
$\mathbf{x}$ does not contain $\varepsilon$.
i do see Roy's point that to mentition $\varepsilon$ here but then ignore
it below is confusing. 
we could put both into a footnote, but we will have to address it later in section 5.1 since we use it there.
here i'm trying to keep but rephrase it. i'm not sure which way is clearer, though.
}
}
$\bm{\pi}$'s score in $\wfsa{A}$ is given by
\begin{align}
\wfsa{A}[\bm{\pi}] = \lambda(q_1) \otimes \left(\bigotimes_{i=1}^{n} \tau(\pi_i) \right) \otimes \rho(q_{n+1}).
\end{align}
\end{definition}
\begin{definition}[string score]
\label{def:master}
Let $\Pi(\seq{x})$ denote the set of all paths in $\wfsa{A}$ that derive $\mathbf{x}$.
Then the score assigned by $\wfsa{A}$ to $\mathbf{x}$ is defined to be
\begin{align}
\label{eq:master}
\wfsa{A}\llbracket \mathbf{x}\rrbracket &= \bigoplus_{\bm{\pi}\in \Pi(\mathbf{x})} \wfsa{A}[\bm{\pi}].
\end{align}
\end{definition}

Because $\K$ is a semiring, $\wfsa{A}\llbracket \seq{x} \rrbracket$ can be computed in time linear in $|\seq{x}|$ by the Forward algorithm~\citep{baum_statistical_1966}. 
Here, for simplicity, we describe the Forward algorithm without $\varepsilon$-transitions.%
\footnote{%
\epstrans s can be handled with a slight modification~\citep{Schwartz:2018}.
Note though that if
$\wfsa{A}$ contains a cycle of \epstrans s, then either $\mathbb{K}$ must follow the \term{star
semiring} laws~\citep{kuich1985semiring}, or the number of consecutive \epstrans s allowed must be capped.}
Its dynamic program is given by:
\begin{subequations}
	\begin{align}
	\label{eq:forward}
	\Omega_0(q) &=
	\lambda(q), \\
	\Omega_{i+1}(q) &=
	 \bigoplus_{q^\prime \in \mathcal{Q}}{ \Omega_{i}(q^\prime) \otimes \tau(q^\prime, q, x_i)}, \\
	\wfsa{A}\llbracket \seq{x}\rrbracket &=
	 \bigoplus_{q \in \mathcal{Q}}{\Omega_n(q) \otimes \rho(q)}.
	\end{align}
\end{subequations}
$\Omega_i(q)$ gives the total score of all paths that derive
$x_1 \ldots x_i$ and end in state $q$.

\begin{example}\label{example:wfsa_b}

Figure~\ref{fig:wfsa:b} diagrams a WFSA $\wfsa{B}$, consisting of two states.
A path starts from the initial state $q_0$ (with $\lambda(q_0) = \bar{1}$);
it then takes any number of ``self-loop'' transitions,
each consuming an input without changing the path score (since it's weighted by $\bar{1}$);
it then consumes an input symbol $\alpha$ and takes a transition weighted by $\mu(\alpha)$,
and reaches the final state $q_1$ (with $\rho(q_1) = \bar{1}$);
it may further consume more input by taking self-loops at $q_1$,
updating the path score by multiplying it by $\phi(\alpha)$ for each symbol $\alpha$.
Then from Definition~\ref{def:master}, we can calculate
that $\wfsa{B}$ gives the empty string score $\bar{0}$, and gives any nonempty
string $\seq{x}=x_1 \dots x_n \in \Sigma^+$ score $\wfsa{B}\llbracket \seq{x} \rrbracket =$
\begin{align}
\label{eq:wfsa:b}
 \bigoplus_{i=1}^{n-1}{\left(
    \mu(x_i)\otimes\bigotimes_{j=i+1}^{n}\phi(x_j)
  \right)}\oplus\mu(x_n).
\end{align}
\end{example}
\noindent
$\wfsa{B}$ can be seen as capturing soft unigram patterns~\citep{davidov2010enhanced},
in the sense that it consumes one input symbol to reach the final state from the initial state.
It is straightforward to design WFSAs capturing longer patterns
by including more states~\citep{Schwartz:2018},
as we will discuss later in \S\ref{sec:relations} and \S\ref{sec:derive}.

\section{Rational Recurrences}
\label{sec:rnn2wfsa}

Before formally defining rational recurrences in \secref{RRNN},
we highlight the connection between WFSAs and RNNs
using a motivating example~(\secref{Motivating}).

\isubsection{A Motivating Example}{Motivating}
We describe a simplified RNN which strips away details of some recent
RNNs, in order to highlight the behaviors of the forget gate and the input.
\begin{example}\label{example:simple}
For an input sequence $\mathbf{x}=x_1 \dots x_n$, let the word embedding vector for $x_t$ be $\mathbf{v}_t$.
As in many gated RNN variants \citep{Hochreiter:1997,Cho:2014}, 
we use a forget gate $\mathbf{f}_t$,
which is computed with an affine transformation followed 
by an elementwise sigmoid function $\bm{\sigma}$.
The current input representation $\mathbf{u}_t$ is similarly computed,
but with an optional nonlinearity (e.g., $\tanh$) $\bm{g}$.
The hidden state 
$\mathbf{c}_t$ can be seen as a weighted sum of 
the previous state and the new input, controlled by the forget gate.
	\begin{subequations}\label{eq:ex}
		\begin{align}
		\mathbf{f}_t&=\bm{\sigma}\bigl(\mathbf{W}_{f}\mathbf{v}_{t} + \mathbf{b}_{f}\bigr),\label{eq:ex:f}\\
		\mathbf{u}_t&=(\mathbf{1}-\mathbf{f}_t)\odot \bm{g}\bigl(\mathbf{W}_{u} \mathbf{v}_{t} + \mathbf{b}_{u}\bigr),\label{eq:ex:u}\\
		\mathbf{c}_t &= \mathbf{f}_t\odot \mathbf{c}_{t-1}+\mathbf{u}_{t}.\label{eq:ex:c}
		\end{align}%
	\end{subequations}%
	The hidden state $\mathbf{c}_t$ can then be used in downstream computation, e.g.,
	to calculate output state $\mathbf{h}_t=\tanh(\mathbf{c}_t)$,
	which is then fed to an MLP classifier.
	We focus only on the recurrent computation.
\end{example}

In Example~\ref{example:simple}, both $\mathbf{f}_t$ and $\mathbf{u}_t$
depend only on the current input token $x_t$ (through $\mathbf{v}_t$), and not
the previous state.
Importantly, the interaction with the previous state $\mathbf{c}_{t-1}$ is \emph{not}
via affine transformations followed by nonlinearities,
as in, e.g., an Elman network~\citep{Elman:1990}, 
where $\mathbf{c}_t = \tanh(\mathbf{W}_c\mathbf{c}_{t-1}+ \mathbf{W}_v\mathbf{v}_{t}+\mathbf{b}_c)$.
As we will discuss later, 
this is important in relating this recurrent update function to WFSAs.

Since the recurrent update in Equation~\ref{eq:ex:c} is elementwise,
for simplicity we focus on just the $i$th dimension.
Unrolling it in time steps, we get
\begin{align}
\label{eq:ex:c:unroll}
\begin{split}
[\mathbf{c}_{t} ]_i
&= [\mathbf{f}_t]_i \, [\mathbf{c}_{t-1}]_i +[\mathbf{u}_{t}]_i\\
&= \sum_{j=1}^{t-1}\left([\mathbf{u}_{j}]_i \prod_{k=j+1}^{t}[\mathbf{f}_{k}]_i\right)
+[\mathbf{u}_{t}]_i,	
\end{split}
\end{align}
where $[\cdot]_i$ denotes the $i$th dimension of a vector. 
As noted by \citet{lee2017ran}, the hidden state at time step $t$ can be seen as
a sum of previous input representations, weighted by the forget gate;
longer histories typically get a smaller weight, since the forget gate values are between 0 and 1 due to the sigmoid function. 

Let's recall the WFSA $\wfsa{B}$ (Figure~\ref{fig:wfsa:b} and Example~\ref{example:wfsa_b})
using the real semiring $\left\langle\mathbb{R}, +, \cdot, 0,1\right\rangle$.
Equation~\ref{eq:ex:c:unroll} is recovered by 
parameterizing $\wfsa{B}$'s weight functions $\mu$ and $\phi$ with
\begin{align}\label{eq:a1:impl}
\begin{array}{ll}
\mu(x_t)= [\mathbf{u}_{t}]_i,& \phi(x_t)= [\mathbf{f}_{t}]_i.
\end{array}
\end{align}
Denote the resulting WFSA by $\wfsa{B}_i$, and we have:

\begin{proposition}\label{prop:ex-wfsa}
	Running a single layer RNN in Example~\ref{example:simple} 
	over any nonempty input string $\mathbf{x}\in \Sigma^+$, 
	the $i$th dimension of its hidden state at time step $t$ 
	equals the score assigned by $\wfsa{B}_i$ to $\mathbf{x}_{:t}$:
	\begin{align}
		[\mathbf{c}_{t} ]_i=\wfsa{B}_i\llbracket \mathbf{x}_{:t}\rrbracket.
	\end{align}
\end{proposition}

In other words, the $i$th dimension of the RNN 
in Example~\ref{example:simple} can be seen as a WFSA structurally equivalent to $\wfsa{B}$.
Its weight functions are implemented as the $i$th dimension of Equations~\ref{eq:ex},
and the learned parameters are the $i$th row of  $\mathbf{W}$ and $\mathbf{b}$.
Then it is straightforward to recover the full $d$-dimensional RNN,
by collecting $d$ such WFSAs, each of which is parametrized 
by a row in the $\mathbf{W}$s and $\mathbf{b}$s.
Based on this observation,
we are now ready to formally define rational recurrences.

\isubsection{Recurrences and Rationality}{RRNN}
For a function $\mathbf{c}\hspace{-.1cm}: \Sigma^\ast\hspace{-.05cm}\to\hspace{-.05cm}\mathbb{K}^d$,
its \textit{recurrence} is said to be the dependence 
of $\mathbf{c}(\mathbf{x}_{:t})$ on $\mathbf{c}(\mathbf{x}_{:t-1})$,
for input sequence $\forall\mathbf{x}\in\Sigma^+$.
We discuss a class of recurrences that can be characterized by WFSAs.
The mathematical counterpart of WFSAs are \textit{rational power series}~\citep{berstel1988rational},
justifying naming such recurrences \textit{rational}:
\begin{definition}[rational recurrence]\label{def:rational}
	The recurrence of $\mathbf{c}\hspace{-.1cm}: \Sigma^\ast\hspace{-.05cm}\to\hspace{-.05cm}\mathbb{K}^d$ is said to be \emph{rational},
	if there exists a set of 
	weighted finite state automata $\{\wfsa{A}_i\}_{i=1}^{d}$
	over alphabet $\Sigma$ and semiring $\left\langle\mathbb{K}, \oplus, \otimes, \bar{0},\bar{1}\right\rangle$
	with both $\oplus$ and $\otimes$ taking constant time and
        space,
	such that $\forall \mathbf{x}\in\Sigma^\ast$,
	\begin{align}
	\bigl[\mathbf{c}(\mathbf{x})\bigr]_i=\wfsa{A}_i\llbracket \mathbf{x}\rrbracket, 
	 \; \forall i \in \{1,2,\dots,d\}.\footnotemark
	\end{align}
\end{definition}
\footnotetext{We restrict that both operations take constant time and space,
	to exclude the use of arbitrarily complex semirings~(\secref{isan}).}
\noindent It directly follows from Proposition~\ref{prop:ex-wfsa} that
\begin{corollary}\label{corollary:ex-rational}
	The recurrence in Example~\ref{example:simple} is rational.
\end{corollary}

\section{Relationship to Existing Neural Models}
\label{sec:relations}
This section studies several recently proposed neural architectures,
and relates them to rational recurrences.
\secref{neural} begins by relating some of them to the RNN defined in Example~\ref{example:simple},
and then to the WFSA $\wfsa{B}$ (Example~\ref{example:wfsa_b}).
We then describe a WFSA similar to $\wfsa{B}$, but with one additional state,
and discuss how it provides a new view of RNN models motivated by $n$-gram features (\secref{ngram}).
In \secref{isan} we study rational recurrences 
that are not elementwise, using an existing model.

In the following discussion, we shall assume the real seimiring, unless otherwise noted.

\subsection{Neural Architectures Related to $\wfsa{B}$}
\label{sec:neural}
Despite its simplicity, Example~\ref{example:simple} corresponds to several existing 
neural architectures. 
For instance, quasi-RNN~(QRNN;~\citealp{Bradbury:2017}) and simple recurrent unit~(SRU;~\citealp{Lei:2017b})
aim to speed up the recurrent computation.
To do so, they drop the matrix multiplication dependence on the previous hidden state, resulting in similar recurrences to that in Example~\ref{example:simple}.%
\footnote{The SRU architecture discussed through this work is based on~\citet{Lei:2017b}.
	In a later updated version,
	\citet{Lei:2018c} introduce diagonal matrix multiplication interaction in the hidden state updates, 
	inspired by~\citep{li2018indrnn}, which yields a recurrence not obviously rational.
}
Other works start from different motivations, but land on similar recurrences,
e.g., strongly-typed RNNs~(T-RNN;~\citealp{Balduzzi:2016}) and its gated variants~(T-LSTM and T-GRU),
and structurally constrained RNNs~(SCRN;~\citealp{mikolov2014learning}).

The analysis in \secref{Motivating} directly applies to SRU, T-RNN, and SCRN.
In fact, Example~\ref{example:simple} presents a slightly more complicated version of them.
In these models, 
input representations are computed without the bias term or any nonlinearity:
~$\mathbf{u}_t=\mathbf{W}_{u}\mathbf{v}_t$.
By Proposition~\ref{prop:ex-wfsa} and Corollary~\ref{corollary:ex-rational}:
\begin{corollary}
The recurrences of single-layer SRU, T-RNN, and SCRN architectures are rational.
\end{corollary}

It is slightly more complicated to analyze
the recurrences of the QRNN, T-LSTM, and T-GRU.
Although their hidden states $\mathbf{c}_t$ 
are updated in the same way as Equation~\ref{eq:ex:c},
the input representations and gates may depend on \textit{previous inputs}.
For example, in T-LSTM and T-GRU, 
the forget gate is a function of two consecutive inputs:
\begin{align}\label{eq:tlstm}
\mathbf{f}_t=\bm{\sigma}\left(\mathbf{V}_f\mathbf{v}_{t-1} + \mathbf{W}_f\mathbf{v}_{t} + \mathbf{b}_f\right).
\end{align}
QRNNs are similar, but may depend on up to $K$ \com{consecutive }tokens,
due to the $K$-window convolutions. 
\citet{eisner_parameter_2002} discuss finite state machines for 
second (or higher) order probabilistic sequence models.
Following the same intuition, we sketch the construction
of WFSAs corresponding to QRNNs with 2-window convolutions in Appendix~\ref{sec:qrnn_proof},
and summarize the key results here:



\begin{proposition}\label{prop:qrnn}
	The recurrences of single-layer T-GRU, T-LSTM, and QRNN are rational.
	In particular, a single-layer $d$-dimensional QRNN using $K$-window convolutions
	can be recovered by a set of $d$ WFSAs, each with $O(2\,\abs{\Sigma}^{K-1})$ states.
\end{proposition}
The size of WFSAs needed to recover QRNN grows exponentially in the window size.
Therefore, at least for QRNNs, Proposition~\ref{prop:qrnn} has
more conceptual value than practical.
\sam{
The ``exponential'' in this last sentence sounds pretty scary.
It's nice that we can construct a WFSA that does the convolution \emph{and} the recurrence all in one...
Couldn't you also consider it two layers though? The first layer does the convolution (a CNN, rational as per SoPa1), and the second layer does the recurrence ($\wfsa{B}$, also rational)?
Would two layers, each simple, be better than one exponentially big layer?
}
\hao{the two-layer argument is a good point, and i think it's  actually doable. 
yet it might be a bit off what we are trying to discuss here:
we somehow assume one-layer architectures in this section, since one can construct crazy stuff if multi-layer architectures are allowed.
we might put a minor point in, e.g., a footnote somewhere. i'll keep thinking about it.}
\subsection{More than Two States}
\label{sec:ngram}
So far our discussion has centered on $\wfsa{B}$, a two-state WFSA capturing unigram patterns~(Example~\ref{example:wfsa_b}).
In the same spirit as going from unigram to $n$-gram features,
one can use WFSAs with more states to capture longer patterns \citep{Schwartz:2018}.
\resolved{
\roy{The relation to n-grams is discussed here for the first time, and at a too high-level. Maybe add a short paragraph to the description of $B$, relating it to unigrams, and hinting at our later analyses with $C$, which corresponds to bigrams?}
\hao{added a paragraph in \S2}
}
In this section we augment $\wfsa{B}$ by introducing more states,
and explore its relationship to some neural architectures motivated by $n$-gram features.
We start with a three-state WFSA as an example, 
and then discuss more general cases.

\begin{figure}[!tb]
	\centering
	\includegraphics[clip,trim=5cm 9.5cm 5cm 9.5cm, width=.7\columnwidth]{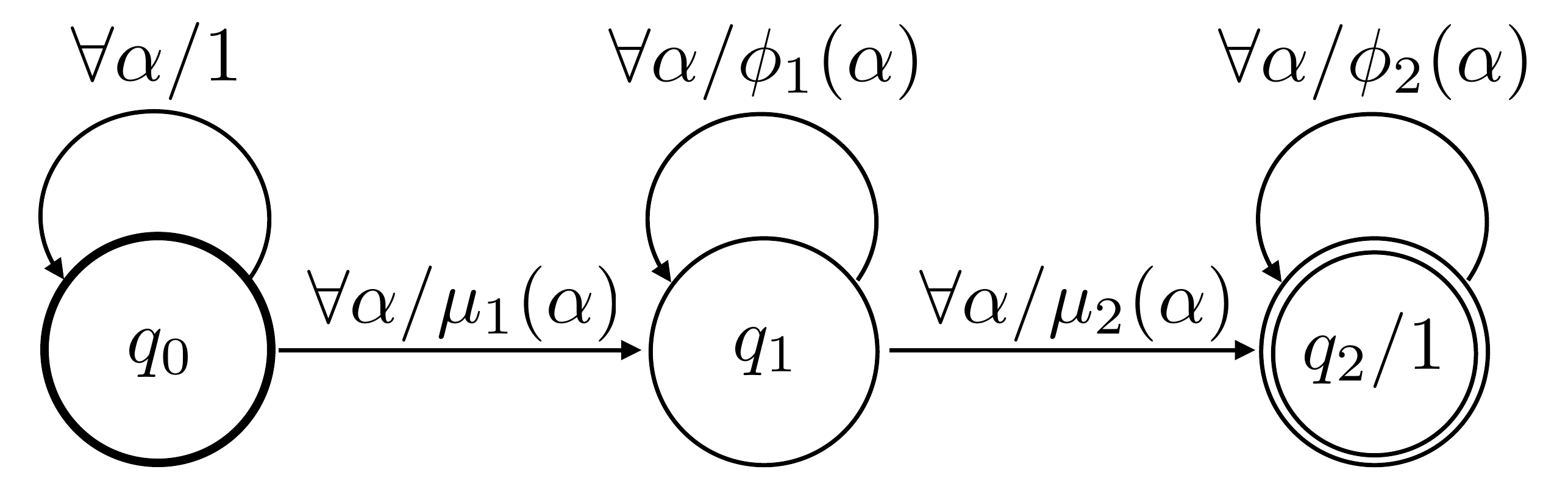}
	\caption{A three-state WFSA $\wfsa{C}$ discussed in \secref{ngram}.}
	\label{fig:wfsa:c}
\end{figure}

Figure~\ref{fig:wfsa:c} diagrams
a WFSA $\wfsa{C}$, augmenting $\wfsa{B}$ with another state.
To reach the final state $q_2$,
at least two transitions must be taken, in contrast to one in $\wfsa{B}$.
History information is decayed by the self-loop at the final state $q_2$, 
assuming $\phi_2$ is between 0 and 1.
$\wfsa{C}$ has another self-loop over $q_1$, weighted by 
$\phi_1\in(0, 1)$.
The motivation is to allow (but down-weight) nonconsecutive bigrams,
as we will soon show.

The scores assigned by
$\wfsa{C}$ can be inductively computed by
applying the Forward algorithm~(\secref{background}).
Given input sequence $\mathbf{x}$ longer than one,
let $\wfsa{C}\llbracket \mathbf{x}_{:0}\rrbracket=0$,
then~$\wfsa{C}\llbracket \mathbf{x}_{:t+1}\rrbracket =$
\begin{align}
	\label{eq:c}
    \wfsa{C}\llbracket \mathbf{x}_{:t}\rrbracket\,\phi_2(x_{t+1}) +\beta_{t}\, \mu_2(x_{t+1}),
\end{align}
where
\begin{align}
\label{eq:c:beta}
\beta_t =\beta_{t-1}\,\phi_1(x_t)+\mu_1(x_t),
\end{align}
and $\beta_0=0$.
Unrolling $\beta_t$ in time, we get $\beta_t =$
\begin{align}
\label{eq:c:beta-unrolled}
\sum_{j=1}^{t-1}\left(\mu_1(x_j) \prod_{k=j+1}^{t}\phi_1(x_k)\right)
+\mu_1(x_t).
\end{align}

Due to the self-loop over state $q_1$,
$\beta_t$ can be seen as a weighted sum of the $\mu_1$ terms up to $x_t$ (Equaltion~\ref{eq:c:beta-unrolled}).
The second product term in Equation~\ref{eq:c}
then provides multiplicative interactions between $\mu_2$,
and the weighted sum of $\mu_1$s.
In this sense, it captures nonconsecutive bigram features.

At a first glance, Equations~\ref{eq:c} and~\ref{eq:c:beta}
resemble recurrent convolutional neural networks~(RCNN;~\citealp{Lei:2016}).
RCNN is inspired by nonconsecutive $n$-gram features and low rank tensor factorization.
It is later studied from a string kernel perspective~\citep{Lei:2017a}.
Here we review its nonlinear bigram version:
\begin{subequations}
	\label{eq:rcnn:c}
	\begin{align}
	\mathbf{c}_{t}^{(1)} &= \mathbf{c}_{t-1}^{(1)} \odot\bm{\lambda}_t  
	+ \mathbf{u}^{(1)}_t,\label{eq:rcnn:c1}\\
	\mathbf{c}_{t}^{(2)} &= \mathbf{c}_{t-1}^{(2)} \odot\bm{\lambda}_t  
	+ \mathbf{c}_{t-1}^{(1)}\odot\mathbf{u}^{(2)}_t,\label{eq:rcnn:c2}
	\end{align}%
\end{subequations}%
where the $\mathbf{u}_t^{(j)}$s are computed similarly to Equation~\ref{eq:ex:u},
and $\mathbf{c}_{t}^{(2)}$ is used as output for onward computation.
Different strategies to computing $\bm{\lambda}_t$
were explored~\citep{Lei:2015,Lei:2016}.
When $\bm{\lambda}_t$ is a constant,
or depends only on $x_t$, 
e.g.,~$\bm{\lambda}_t=\bm{\sigma}(\mathbf{W}_{\lambda}\mathbf{v}_t + \mathbf{b}_{\lambda})$,
the $i$th dimension of Equations~\ref{eq:rcnn:c} can be recovered from Equation~\ref{eq:c}, by letting
\begin{align}\label{eq:c:impl}
\mu_j(x_t)= [\mathbf{u}_{t}^{(j)}]_i,\, \phi_j(x_t)= [\bm{\lambda}_{t}]_i, \; j=1,2.
\end{align}
It is straightforward to generalize the above discussion to higher order cases:
$n$-gram RCNN corresponds to WFSAs with $n+1$ states,
constructed similarly to how we build $\wfsa{C}$ from $\wfsa{B}$ (Appendix~\ref{sec:rcnn_proof}).
\begin{proposition}\label{proposition:rcnn}
For a single-layer RCNN with $\bm{\lambda}_t$ being a constant
or depending only on $x_t$, the recurrence is rational.
\end{proposition}
\noindent
As noted later in \S\ref{sec:isan},
its recurrence may not be rational
when~$\bm{\lambda}_t=\bm{\sigma}(\mathbf{W}_{c}\mathbf{c}_{t-1} + \mathbf{W}_{\lambda}\mathbf{v}_t +\mathbf{b}_{\lambda})$.

\subsection{Beyond Elementwise Operations}
\label{sec:isan}
So far we have discussed rational recurrences for models using 
elementwise recurrent updates (e.g., Equation~\ref{eq:ex:c}).
This section uses an existing model as an example,
to study a rational recurrence that is not elementwise.
We focus on the input switched affine network (ISAN;~\citealp{Foerster:2017}).
Aiming for efficiency and interpretability,
it does not use any explicit nonlinearity;
its affine transformation parameters depend only on the input:
\begin{align}\label{eq:isan}
\mathbf{c}_t = \mathbf{W}_{x_t}\mathbf{c}_{t-1}  +\mathbf{b}_{x_t}.
\end{align}
Due to the matrix multiplication,
the recurrence of a single-layer ISAN is not elementwise.
Yet, we argue that it is rational.
We will sketch the proof for a 2-dimensional case,
and it is straightforward to generalize to higher dimensions (\appref{isan_proof}).

\begin{figure}[!t]
	\centering
	\includegraphics[clip,trim=1cm 9cm 1cm 10cm, width=0.9\columnwidth]{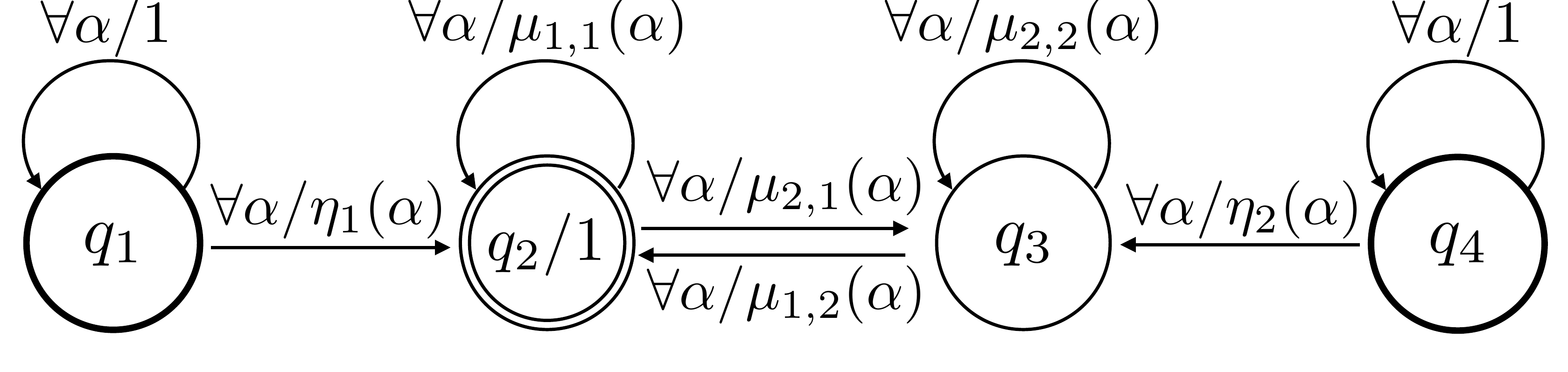}
	\caption{WFSA $\wfsa{D}_1$ discussed in \secref{isan}.
		Two initial states $q_1$ and $q_4$ are used here.}
	\label{fig:wfsa:d1}
\end{figure}

We define two WFSAs, each recovering one dimension of ISAN's recurrent updates.
Figure~\ref{fig:wfsa:d1} diagrams one of them, $\wfsa{D}_1$.
The other one, 
$\wfsa{D}_2$, is identical (including shared weights), except using $q_3$ instead of $q_2$ as the final state.
For any nonempty input sequence~$\mathbf{x}\in\Sigma^+$,
the scores assigned by $\wfsa{D}_1$ and $\wfsa{D}_2$ can be inductively computed by
applying the Forward algorithm.
Letting $\wfsa{D}_1\llbracket \mathbf{x}_{:0}\rrbracket=\wfsa{D}_2\llbracket \mathbf{x}_{:0}\rrbracket=0$,
for $t\geq 1$
\begin{align}
\begin{split}
\begin{bmatrix}
\wfsa{D}_1\llbracket \mathbf{x}_{:t}\rrbracket\\
\wfsa{D}_2\llbracket \mathbf{x}_{:t}\rrbracket  
\end{bmatrix}
=
\widetilde{\mathbf{W}}_{x_t} 
\begin{bmatrix}
\wfsa{D}_1\llbracket \mathbf{x}_{:t-1}\rrbracket\\
\wfsa{D}_2\llbracket \mathbf{x}_{:t-1}\rrbracket  
\end{bmatrix}
+\widetilde{\mathbf{b}}_{x_t},
\end{split}
\end{align}
where
\begin{align}
\begin{split}
\widetilde{\mathbf{W}}_{x_t} 
&=
\begin{bmatrix}
\mu_{1,1}(x_t) & \mu_{1,2}(x_t)\\
\mu_{2,1}(x_t) & \mu_{2,2}(x_t)   
\end{bmatrix},\\
\widetilde{\mathbf{b}}_{x_t}
& =
\begin{bmatrix}
\eta_1(x_t)\\
\eta_2(x_t)
\end{bmatrix}.
\end{split}
\end{align}
Then Equation~\ref{eq:isan}, in the case of hidden size 2, is recovered by
letting $\mathbf{W}_{x_t}=\widetilde{\mathbf{W}}_{x_t}$ and $\mathbf{b}_{x_t}=\widetilde{\mathbf{b}}_{x_t}$.

\begin{proposition}\label{proposition:isan}
	The recurrence of a single-layer ISAN is rational.
\end{proposition}

\begin{corollary}\label{corollary:elman}
	For a single-layer Elman network,
	in the absence of any nonlinearity,
	the recurrence is rational.
\end{corollary}

\paragraph{Discussion.}
It is known that an Elman network can approximate 
any recursively computable partial function~\citep{siegelmann1995rnn}.
On the other hand, in their single-layer cases,
WFSAs (and thus models with rational recurrences) 
are restricted to rational series~\citep{schutzenberger1961fsa}.
Therefore, we hypothesize that models like Elman networks, LSTMs, and GRUs,
where the recurrences depend on previous states through affine transformations 
followed by nonlinearities, are not rational.

This work does not intend to propose rational recurrences as a
concept general enough to include most existing RNNs.
Rather, we wish to study a more constrained class of methods
to better understand the connections between WFSAs and RNNs.
Therefore in Definition~\ref{def:rational}, we restrict 
the semirings to be ``simple,''
in the sense that both operations take constant time and space.
Such a restriction aims to exclude the possibility of hiding arbitrarily
complex computations inside the semiring,
which might allow RNNs to satisfy the definition in a trivial and unilluminating way.

Such theoretical limitations might be less severe than they appear,
since it is not yet entirely clear what they correspond to in practice, 
especially when multiple vertical layers of these models are used~\citep{leshno1993multilayer}.
We defer to future work the further study of the connections between WFSAs and Elman-style RNNs. 

Closing this section,
Table~\ref{tab:wfsa} summarizes the discussed recurrent neural architectures and their corresponding WFSAs.

\begin{table}[tb]
	\setlength{\tabcolsep}{5pt}
	\centering
	\small
	\begin{tabularx}{0.48\textwidth}{@{} l@{\,\,}l @{\;} p{4.0cm} @{} c @{}} 
		
		\toprule
		
		&\textbf{Models} & \textbf{Recurrence Function} & \textbf{WFSA}  \\
		
		\midrule
		 \multirow{ 2}{*}{\S\ref{sec:neural}} & 
		SRU, SCRN & \multirow{ 2}{*}{$\mathbf{c}_t = \mathbf{f}_t\odot \mathbf{c}_{t-1}+\mathbf{u}_{t}$} 
		& \multirow{ 2}{*}{$\wfsa{B}$}  \\
		& T-RNN, QRNN && \\
		
		\midrule
		
		\S\ref{sec:ngram} &
		RCNN & ${\begin{aligned}
		\mathbf{c}_{t}^{(1)}\hspace{-.1cm}&= \mathbf{c}_{t-1}^{(1)} \odot\bm{\lambda}_t  
		+ \mathbf{u}^{(1)}_t\\
		\mathbf{c}_{t}^{(2)}\hspace{-.1cm}&= \mathbf{c}_{t-1}^{(2)} \odot\bm{\lambda}_t  
		+ \mathbf{c}_{t-1}^{(1)}\odot\mathbf{u}^{(2)}_t \end{aligned}}$
		 & $\wfsa{C}$  \\
		 
		\midrule

		\S\ref{sec:isan} & 
		ISAN & $\mathbf{c}_t = \mathbf{W}_{x_t}\mathbf{c}_{t-1}  +\mathbf{b}_{x_t}$ & $\wfsa{D}_1, \wfsa{D}_2$  \\ 
		\bottomrule
	\end{tabularx}
	\caption{Recurrent neural network architectures discussed in \S\ref{sec:relations}
	and their corresponding WFSAs.
	\S\ref{sec:neural}: SRU~\citep{Lei:2017b},
	SCRN~\citep{mikolov2014learning},
	T-RNN and its gated variants~\citep{Balduzzi:2016},
	and QRNN~\citep{Bradbury:2017};
	\S\ref{sec:ngram}: RCNN~\citep{Lei:2016};
	\S\ref{sec:isan}: ISAN~\citep{Foerster:2017}.}
	\label{tab:wfsa}
\end{table}

\section{Deriving Neural Models from WFSAs}
\label{sec:derive}

Rational recurrences provide a new view of several recently proposed neural models.
Based on such observations, this section aims to explore 
potential approaches to designing neural architectures
in a more interpretable and intuitive way: by deriving them from WFSAs.
\secref{unigram_bigram} studies an interpolation of unigram and bigram features by
combining 2-state and 3-state WFSAs (Figures~\ref{fig:wfsa:b} and~\ref{fig:wfsa:c}).
We then explore alternative semirings (\secref{semiring}), 
an approach orthogonal to what we've discussed so far.

We note that our goal is not to devise new state-of-the-art architectures.
Rather, we illustrate a new design process for neural architectures that
draws inspiration from WFSAs.
That said, in our experiments (\secref{experiments}), 
one of our new architectures performs as well as or better than strong baselines.

\subsection{Aggregating Different Length Patterns}
\label{sec:unigram_bigram}

We start by presenting a straightforward extension to 2-state and 3-state rational models: 
one combining both.
It is inspired by many classical NLP models, 
where unigram features and higher-order ones are interpolated.

\begin{figure}[!t]
	\centering
	\includegraphics[clip,trim=4cm 7cm 4cm 7cm, width=.7\columnwidth]{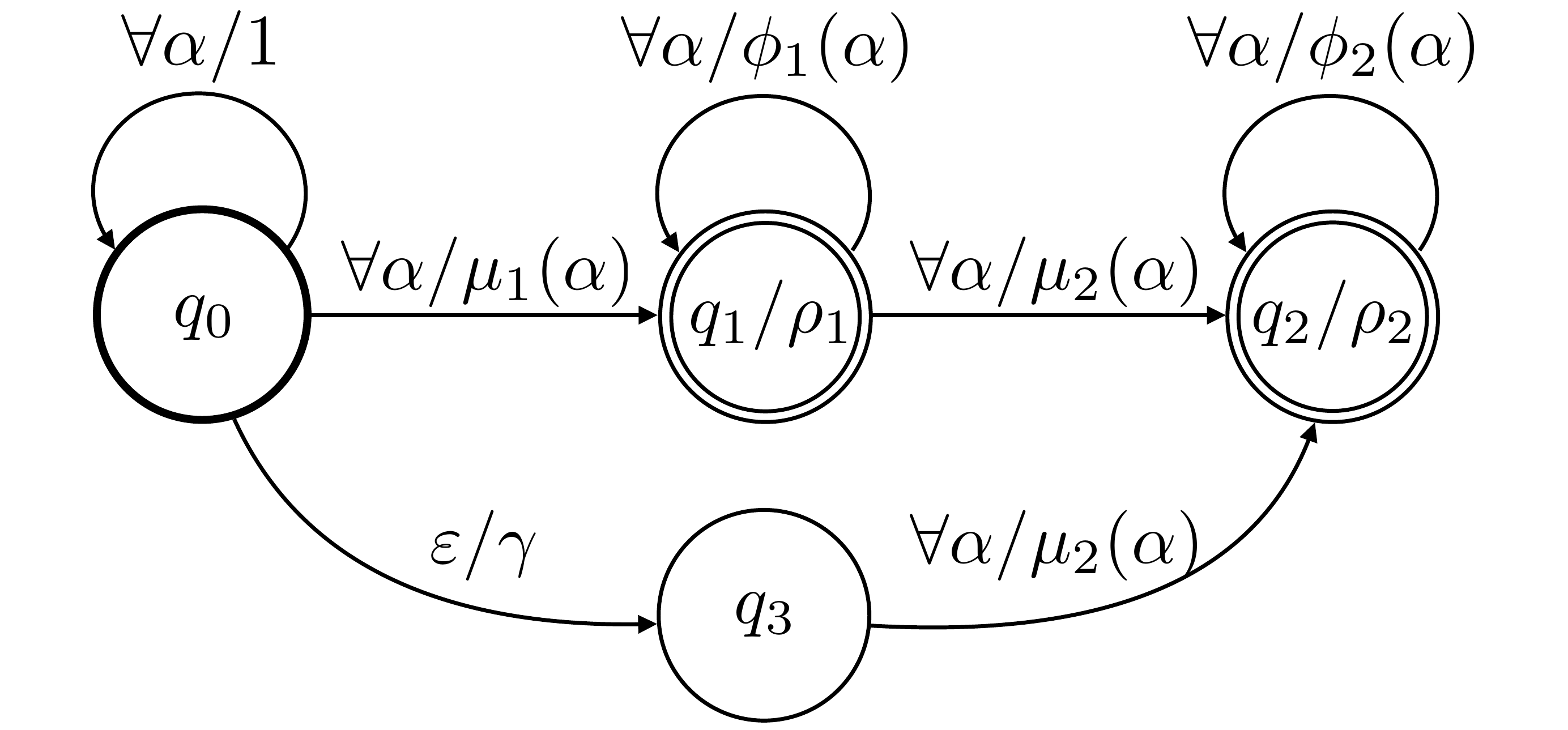}
	\caption{A WFSA $\wfsa{F}$ that combines both unigram and bigram features~(\secref{unigram_bigram}).
		Two final states $q_1$ and $q_2$ are used, with weights $\rho_1$ and $\rho_2$, respectively.}
	\label{fig:wfsa:e}
\end{figure}

Figure~\ref{fig:wfsa:e} diagrams a 4-state WFSA $\wfsa{F}$.
Compared to $\wfsa{C}$ (Figure~\ref{fig:wfsa:c}),
$\wfsa{F}$ uses $q_1$ as a second final state,
aiming to capture both unigram and bigram patterns,
since a path is allowed to stop at $q_1$ after consuming one input.
The final states are weighted by $\rho_1$ and $\rho_2$ respectively.
Another notable modification is the additional state $q_3$,
which is used to create a ``shortcut'' to reach $q_2$,
together with an $\varepsilon$-transition.
Specifically, starting from $q_0$, 
a path can now take the $\varepsilon$-transition and reach $q_3$,
and then take a transition with weight $\mu_2$ to reach $q_2$.
Recall from \S\ref{sec:background}, that $\varepsilon$-transitions
do not consume any input,
yet they can still be weighted by a (parameterized) function $\gamma$ 
\textit{not} depending on the inputs.
The $\varepsilon$-transition allows for skipping the 
first word in a bigram.
It can be discouraged by using $\gamma\in(0,1)$, just as we do in our experiments.

\paragraph{Deriving the neural architecture.}
As in \S\ref{sec:rnn2wfsa}, we relate hidden states of an RNN
to the scores assigned by WFSAs to input strings.
We then derive the neural architecture
\com{in a mechanical way }with a dynamic program.
Here we keep the discussion self-contained
by explicitly overviewing the procedure.
It is a direct application of 
 the Forward algorithm 
(\S\ref{sec:background}), though now
in a form that deals with the $\varepsilon$-transition.
Such an approach applies,  of course, to more general cases,
as noted by \citet{Schwartz:2018}. 

Given an input string $\mathbf{x}\in\Sigma^+$,
let $z^{(j)}_t$ denote the total score of all paths
landing in state $q_j$ \textit{just} after consuming $x_t$.
Let $z^{(j)}_0 = 0$, then for $t\geq1$,
\begin{align*}	
\begin{split}
	z^{(0)}_t &= 1\\
	z^{(1)}_t &= z^{(1)}_{t-1}\,\phi_1(x_t) +z^{(0)}_{t-1}\,\mu_1(x_t) \\
	z^{(3)}_t &= z^{(0)}_{t}\,\gamma \\
	z^{(2)}_t &= z^{(2)}_{t-1}\,\phi_2(x_t)
 + (z^{(1)}_{t-1}+z^{(3)}_{t-1})\, \mu_2(x_t)\\
	\wfsa{F}\llbracket \mathbf{x}_{:t}\rrbracket
	&=\rho_1\,z^{(1)}_t + \rho_2\,z^{(2)}_t.
\end{split}
\end{align*}
We now collect $d$ of these WFSAs to construct an RNN,
and we parameterize their weight functions with
the technique we've been using:
\begin{subequations}\label{eq:both}
	\begin{align}
	\mathbf{c}^{(1)}_t &= \mathbf{c}^{(1)}_{t-1}\odot\mathbf{f}_t^{(1)} + \mathbf{u}_t^{(1)}, \\
	\mathbf{c}^{(2)}_t
	&=\mathbf{c}^{(2)}_{t-1}\odot\mathbf{f}^{(2)}_t 
	+ (\mathbf{c}^{(1)}_{t-1}+\mathbf{r})\odot \mathbf{u}_t^{(2)},\\
	\mathbf{c}_t
	&=\mathbf{p}^{(1)}\odot \mathbf{c}^{(1)}_t + \mathbf{p}^{(2)}\odot\mathbf{c}^{(2)}_t,\label{eq:unigram_bigram:c}
	\end{align}
\end{subequations}
where
\begin{subequations}\label{eq:bothb}
\begin{align}
\mathbf{f}_t^{(j)}&=\bm{\sigma}\bigl(\mathbf{W}_{f}^{(j)}\mathbf{v}_{t} + \mathbf{b}_{f}^{(j)}\bigr),\\
\mathbf{u}_t^{(j)}&=(\mathbf{1}-\mathbf{f}_t^{(j)})\odot \bm{g}\bigl(\mathbf{W}_{u}^{(j)} \mathbf{v}_{t} + \mathbf{b}_{u}^{(j)}\bigr),\\
&\; \mathbf{p}^{(j)}=\bm{\sigma}(\mathbf{b}_{p}^{(j)}), \quad \mathbf{r}=\bm{\sigma}(\mathbf{b}_{r}).
\end{align}
\end{subequations}
The $\mathbf{p}$ vectors correspond to the final state weights $\rho_1$ and $\rho_2$.
Despite the similarities,
$\mathbf{p}$ are different from output gates~\citep{Bradbury:2017},
since the former do not depend on the input, and are parameterized (through a sigmoid) by two leanred vectors $\mathbf{b}_{p}^{(j)}$.
The same applies to $\mathbf{r}$ and $\mathbf{b}_{r}$, which correspond to the weights for $\varepsilon$-transitions $\gamma$.


\subsection{Alternative Semirings}
\label{sec:semiring}
Our new understanding of \names allows us to consider a different kind
of extension: replacing the semiring.
We introduce an example, which modifies Example~\ref{example:simple}
by replacing its real (plus-times) semiring with the max-plus semiring~$\left\langle\mathbb{R}\cup\{-\infty\}, \max, +, -\infty,0\right\rangle$:

\begin{example}\label{example:maxplus}
	\begin{subequations}
		\label{eq:maxplus}
		\begin{align}
		\mathbf{f}_t&=\log\bm{\sigma}\bigl(\mathbf{W}_{f}\mathbf{v}_{t} + \mathbf{b}_{f}\bigr),\label{eq:maxplus:f}\\
		\mathbf{u}_t&=\bm{g}\bigl(\mathbf{W}_{u} \mathbf{v}_{t} + \mathbf{b}_{u}\bigr),\label{eq:maxplus:u}\\
		\mathbf{c}_t &= \max\{\mathbf{f}_t + \mathbf{c}_{t-1}, \mathbf{u}_{t}\}.\label{eq:maxplus:c}
		\end{align}
	\end{subequations}
\end{example}
\noindent 
Example~\ref{example:maxplus} does not use the forget gate
when computing $\mathbf{u}_t$~(Equation~\ref{eq:maxplus:u}), 
which is different from its plus-times counterpart, 
where~$\mathbf{u}_t=(\mathbf{1}-\mathbf{f}_t)\odot \bm{g}\bigl(\mathbf{W}_{u} \mathbf{v}_{t} + \mathbf{b}_{u}\bigr)$.
The reason is that,
unlike the real semiring,
the max-plus semiring lacks a well-defined negation.
Possible alternatives include taking the $\log$ of a separate input gate, 
or using $\log(\mathbf{1}-\mathbf{f}_t)$,
which we leave for future work.

Example~\ref{example:maxplus} can be seen as 
replacing sum-pooling with max-pooling.
Both max and sum-pooling have been used successfully in vision and NLP models.
Intuitively, max-pooling ``detects'' the occurrence of a pattern while sum-pooling ``counts'' the occurrence of a pattern. 
One advantage of max operator is that the model's decisions can be back-traced and interpreted, as argued by \citet{Schwartz:2018}. 
Such a technique is applicable to all the models with rational recurrences.

\section{Experiments}
\label{sec:experiments}
\begin{table}[tb]
	\setlength{\tabcolsep}{5pt}
	\centering
	\begin{tabularx}{0.48\textwidth}{@{} l c c c@{}} 
		
		\toprule
		
		\textbf{Model} & \textbf{Unigram} & \textbf{Bigram} &\textbf{Semiring}  \\
		\midrule
		\unigram & \checkmark & &real\\
		\unimp & \checkmark & & max-plus\\
		\bigram & & \checkmark &real\\
		\both &\checkmark & \checkmark &real\\
		
		\bottomrule
	\end{tabularx}
	\caption{Rational recurrent neural architectures compared in the experiments~(\secref{models}).}
	\label{tab:wfsa_baseline}
\end{table}

This section\com{ empirically} evaluates four rational RNNs
on language modeling~(\S\ref{subsec:lm}) and text categorization~(\S\ref{subsec:textcat}).
Our goal is to compare the behaviors of models derived from different WFSAs,
showing that our understanding of WFSAs allows us to improve existing rational models.

\isubsection{Compared Models}{models}
Our comparisons focus on the \textit{recurrences} of the models,
i.e., how the hidden states $\mathbf{c}_t$ are computed~(e.g., 
Equations~\ref{eq:ex:c} and \ref{eq:unigram_bigram:c}).
Therefore we follow~\citet{Lei:2017b}
and use $\mathbf{u}_t^{(j)}=\mathbf{W}_u^{(j)}\mathbf{v}_t^{(j)}$
across all compared models, listed below and as well as in Table~\ref{tab:wfsa_baseline}:
\begin{compactitem}[$\bullet$]
	\item \unigram, with real semiring (\secref{neural});		 
	\item \unimp, with max-plus semiring~(\secref{semiring});
	\item \bigram, with real semiring  (\secref{ngram});		  	
	\item \both, with real semiring (\secref{unigram_bigram}).		  	
\end{compactitem}		  

We also compare to an LSTM baseline.
Aiming to control for comfounding factors,
we do not use highway connections in any of the models.%
\footnote{Thus \unigram~ is essentially an SRU without highway connections.
We denote it differently, to note its differences from the original implementation \cite{Lei:2017b}.
Similarly, we do not denote \bigram as RCNN~\citep{Lei:2016}.}
In the interest of space, 
the full architectures and hyperparameters are detailed in Appendices~\ref{sec:model_details} and \ref{sec:Experimental}.

\subsection{Language Modeling}
\label{subsec:lm}
\paragraph{Dataset and implementation.}
We experiment with the Penn Treebank corpus~(PTB;~\citealp{Marcus:1993}).
We use the preprocessing and splits from~\citet{mikolov2010rnn},
resulting in a vocabulary size of 10K and 1M tokens.

Following standard practice, we treat the training data as one long sequence,
split into mini batches, and train using 
BPTT truncated to 35 time steps~\citep{Williams:1990}.
The input embeddings and output softmax weights are tied~\citep{press2017using}.

\paragraph{Results.}

Following \citet{collins2016capacity} and \citet{melis2017sota},
we compare models controlling for parameter budget.
\tabref{lm_results} summarizes language modeling perplexities on PTB test set.
The middle block compares all models with two layers and 10M trainable parameters.
\unigram~and \bigram~achieve roughly the same performance;
interpolating both unigram and bigram features,
\both~outperforms others by more than 2.9 test perplexity.
For the three-layer and 24M setting (the bottom block),
we observe similar trends,
except that~\bigram slightly underperforms~\unigram.
Here~\both~outperforms others by more than 2.1 perplexity.

\begin{table}[tb]
	\centering
	\begin{tabulary}{0.47\textwidth}{@{}l  cc cc@{}} 
		
		\toprule
		
		\textbf{Model}
		& $\boldsymbol{\ell}$
		& \textbf{\# Params.}
		& \textbf{Dev.}
		& \textbf{Test}\\
		
		\midrule
		
		\lstm & 2 & 24M & 73.3 & 71.4\\
		\lstm & 3 & 24M & 78.8 & 76.2\\
		\midrule[.03em]
		\unigram & 2 & 10M & 73.1 & 69.2\\
		\unimp & 2 & 10M & 75.1& 71.7 \\
		\bigram &  2 & 10M & 72.5 & 69.5\\
		\both & 2 & 10M & {\bf 69.5} & {\bf 66.3}\\ 
		
		\midrule[.03em]
		
		\unigram & 3 & 24M & 68.7& 65.2 \\
		\unimp & 3 & 24M & 70.8& 66.9 \\
		\bigram &  3 & 24M & 70.0 &67.0 \\
		\both & 3 & 24M &{\bf 66.0} & {\bf 63.1} \\ 
		
		\bottomrule

	\end{tabulary}
	\caption{\label{tab:lm_results} Language modeling perplexity  on PTB test set (lower is better).
		\lstm~numbers are taken
		from~\citet{Lei:2017b}.  $\ell$ denotes the number of
		layers.
		Bold font indicates best performance.}
	\label{lm}
\end{table}

Using a max-plus semiring,
~\unimp~\emph{underperforms}~\unigram~under both settings.
Possible reasons could be the suboptimal design choice 
for computing input representations in the former~(\secref{semiring}).
Finally, most compared models outperform the \lstm~baselines,
whose numbers are taken from~\citet{Lei:2017b}.\footnote{\citet{melis2017sota} point out that carefully tuning LSTMs
can achieve much stronger performance, at the cost of exceptionally large amounts of computational resources for tuning.\com{, . Yet we note that none of the works compared in \tabref{lm_results}
has access to the amount of computational resources needed for
hyperparameter tuning as in~\citet{melis2017sota}}}


\subsection{Text Classification}
\label{subsec:textcat}
\paragraph{Implementation.}
We use unidirectional 2-layer architectures for all compared models.
To build the classifiers, we feed the final RNN hidden states into a 2-layer $\tanh$-MLP. 
Further implementation details are described in Appendix~\ref{sec:Experimental}.

\paragraph{Datasets.}
We experiment with four binary text classification datasets, described below.
\begin{table}[tb]
	\setlength{\tabcolsep}{5pt}
	\centering
	\begin{tabulary}{0.47\textwidth}{@{}l cccc@{}} 
		
		\toprule
		
		\textbf{Split} & \textbf{Amazon} & \textbf{SST} & \textbf{subj} & \textbf{CR}\\
		
		\midrule
		Train & 20K & 6.9K & 8K & 3.0K \\
		Dev. &{\color{white}{0}}5K & 0.9K & 1K & 0.4K\\
		Test & 25K & 1.8K & 1K & 0.4K\\ 
		
		\bottomrule
		
	\end{tabulary}
	\caption{Number of instances in the text classification datasets~(\S\ref{subsec:textcat}).}
	\label{tab:textcat_data}
\end{table}


\begin{compactitem}[$\bullet$]
	\item\textbf{Amazon} (electronic product review corpus; \citealp{McAuley:2013}).\camready{\footnote{\url{http://riejohnson.com/cnn_data.html}}}
	We focus on the positive and negative reviews.
	\item\textbf{SST}
	(Stanford sentiment treebank;~\citealp{Socher:2013}).\camready{\footnote{\url{nlp.stanford.edu/sentiment/index.html}}}
	We focus on the binary classification task. 
	SST provides labels for syntactic phrases;
	we experiment with a more realistic setup,
	and consider only complete sentences at either training or evaluating time.
	\item\textbf{subj} 
	(Subjectivity dataset; \citealp{Pang:2004}).
	As {\bf subj} doesn't come with official splits, we randomly split it
	to train (80\%), development (10\%), and test (10\%) sets.
	\item\textbf{CR}
	(customer reviews dataset; \citealp{Hu:2004}).%
	\camready{\footnote{\url{http://www.cs.uic.edu/?liub/FBS/sentiment-analysis.html}}}
	As with {\bf subj}, we randomly split this dataset using the same ratio.
\end{compactitem}
\tabref{textcat_data} summarizes the sizes of the datasets.

\paragraph{Results.}
\tabref{textcat} summarizes text classification test accuracy. 
We report the average performance of 5 trials different only in random seeds. 
\both~outperforms all other models on 3 out of the 4 datasets.
For Amazon, the largest one,
we do not observe significant differences between \both~and \bigram,
while both outperform others.
This may suggest that 
the interpolation of unigram and bigram features by \both~
is especially useful in small data setups.
As in the language modeling experiments,
 \unimp underperforms all other models in most cases, and in
 particular \unigram.
These results provide evidence that replacing the real
semiring 
in rational models might be challenging. 
We leave further exploration to future work.

\begin{table}[tb]
\setlength{\tabcolsep}{2.1pt}
	\centering
	\begin{tabulary}{0.47\textwidth}{@{}l cccc@{}} 
		
		\toprule
		
		\textbf{Model} & \textbf{Amazon} & \textbf{SST} & \textbf{subj} & \textbf{CR}\\
		
		\midrule
		
		\lstm & 91.2$_{\pm 0.3}$ &85.1$_{\pm  0.6}$&93.3$_{\pm  0.6}$& 82.4$_{\pm  1.5}$\\
		\midrule[.03em]
		\unigram & 92.4$_{\pm  0.1}$ & 85.8$_{\pm 	0.3  }$ & 93.9$_{\pm 0.4}$ & 84.1$_{\pm  1.0}$\\
	 	\bigram & {\bf 92.8}$_{\pm 0.2}$ & 84.8$_{\pm  0.4}$ & 93.8$_{\pm  0.6}$ & 84.5$_{\pm  0.9}$\\ 
	 	
	 	\midrule[.03em]
		\unimp & 89.2$_{\pm 3.1}$& 84.9$_{\pm 0.4 }$ & 92.6$_{\pm 0.5}$ & 84.3$_{\pm 0.5}$\\	 	
		\both & 92.7$_{\pm  0.2}$ & {\bf 86.5}$_{\pm  0.6}$ & {\bf 94.8}$_{\pm  0.5}$ & {\bf 85.1}$_{\pm  0.5}$ \\
		
		\bottomrule
		
		\end{tabulary}
	\caption{Text classification test accuracy averaged over 5 runs. 
		$\pm$ denotes standard deviation, 
		and bold font indicates best averaged performance.}
	\label{tab:textcat}
\end{table}

\section{Related Work}
\label{sec:related}
\paragraph{Weighted finite state automata.}
WFSAs were once popular among many sequential tasks~\interalia{mohri2002weighted,kumar2003weigthed,cortes_rational_2004,pardo2005modeling,moore_juicer_2006},
and are still successful in morphology~\interalia{dreyer_non-parametric_2011,cotterell_modeling_2015,rastogi_weighting_2016}.
Compared to neural networks, WFSAs are better understood theoretically
and arguably more interpretable.
They were recently revisited in combination with the former
in, e.g., text generation~\citep{ghazvininejad2016generating,ghazvininejad2017hafez,lin2017adversarial}
and automatic music accompaniment~\citep{forsyth2016automatic}.

\paragraph{Recurrent neural networks.}
RNNs~\citep{Elman:1990,jordan1989serial} 
prove to be strong models for sequential data~\citep{siegelmann1995rnn}.
Besides the perhaps most notable gated variants~\citep{Hochreiter:1997,Cho:2014},
extensive efforts have been devoted to developing alternatives~\interalia{Balduzzi:2016,miao2016simplifying,zoph2017neural,lee2017ran,Lei:2017a,vaswani2017attention,Gehring:2017}.
Departing from the above approaches,
this work derives RNN architectures drawing inspiration from WFSAs.

Another line of work studied the connections between WFSAs and RNNs in terms of modeling capacity,
both
empirically~\interalia{kolen1993fools,Giles:1992,weiss2018practical}
and
theoretically~\interalia{Cleeremans:1989,Visser:2001,Chen:2018}.
\section{Conclusion}
\label{sec:conclusion}
We presented \emph{rational recurrences},
a new construction to study the recurrent updates in RNNs,
drawing inspiration from WFSAs. 
We showed that rational recurrences are in frequent use by several
recently proposed recurrent neural architectures,
providing new understanding of them.
Based on such connections,
we discussed approaches to deriving novel neural architectures from WFSAs.
Our empirical results demonstrate the potential of doing so.
We publicly release our implementation at~\repo.

\section*{Acknowledgments}
We thank Jason Eisner, Luheng He, Tao Lei, Omer Levy, members of the ARK lab at the University of Washington, 
and researchers at the Allen Institute for Artificial Intelligence for their helpful comments on an early version of this work,
and the anonymous reviewers for their valuable feedback.
We also thank members of the Aristo team at the Allen Institute for Artificial Intelligence for their support with the Beaker experimentation system.
This work was supported in part by NSF grant IIS-1562364 and by the NVIDIA Corporation through the donation of a Tesla GPU.

\clearpage
\bibliography{emnlp2018}

\begin{thebibliography}{66}
\expandafter\ifx\csname natexlab\endcsname\relax\def\natexlab#1{#1}\fi

\bibitem[{Bahdanau et~al.(2015)Bahdanau, Cho, and Bengio}]{bahdanau2014neural}
Dzmitry Bahdanau, Kyunghyun Cho, and Yoshua Bengio. 2015.
\newblock Neural machine translation by jointly learning to align and
  translate.
\newblock In \emph{Proc. of ICLR}.

\bibitem[{Balduzzi and Ghifary(2016)}]{Balduzzi:2016}
David Balduzzi and Muhammad Ghifary. 2016.
\newblock Strongly-typed recurrent neural networks.
\newblock In \emph{Proc. of ICML}.

\bibitem[{Baum and Petrie(1966)}]{baum_statistical_1966}
Leonard~E. Baum and Ted Petrie. 1966.
\newblock Statistical inference for probabilistic functions of finite state
  {Markov} chains.
\newblock \emph{The Annals of Mathematical Statistics}, 37(6):1554--1563.

\bibitem[{Berstel and Reutenauer(1988)}]{berstel1988rational}
Jean Berstel, Jr. and Christophe Reutenauer. 1988.
\newblock \emph{Rational Series and Their Languages}.
\newblock Springer-Verlag, Berlin, Heidelberg.

\bibitem[{Bradbury et~al.(2017)Bradbury, Merity, Xiong, and
  Socher}]{Bradbury:2017}
James Bradbury, Stephen Merity, Caiming Xiong, and Richard Socher. 2017.
\newblock Quasi-recurrent neural network.
\newblock In \emph{Proc. of ICLR}.

\bibitem[{Chen et~al.(2018)Chen, Gilroy, Knight, and May}]{Chen:2018}
Yining Chen, Sorcha Gilroy, Kevin Knight, and Jonathan May. 2018.
\newblock Recurrent neural networks as weighted language recognizers.
\newblock In \emph{Proc. of NAACL}.

\bibitem[{Cho et~al.(2014)Cho, Van~Merri{\"e}nboer, Gulcehre, Bahdanau,
  Bougares, Schwenk, and Bengio}]{Cho:2014}
Kyunghyun Cho, Bart Van~Merri{\"e}nboer, Caglar Gulcehre, Dzmitry Bahdanau,
  Fethi Bougares, Holger Schwenk, and Yoshua Bengio. 2014.
\newblock Learning phrase representations using {RNN} encoder-decoder for
  statistical machine translation.
\newblock In \emph{Proc. of EMNLP}.

\bibitem[{Cleeremans et~al.(1989)Cleeremans, Servan-Schreiber, and
  McClelland}]{Cleeremans:1989}
Axel Cleeremans, David Servan-Schreiber, and James~L. McClelland. 1989.
\newblock Finite state automata and simple recurrent networks.
\newblock \emph{Neural computation}, 1(3):372--381.

\bibitem[{Collins et~al.(2017)Collins, Sohl-Dickstein, and
  Sussillo}]{collins2016capacity}
Jasmine Collins, Jascha Sohl-Dickstein, and David Sussillo. 2017.
\newblock Capacity and trainability in recurrent neural networks.
\newblock In \emph{Proc. of ICLR}.

\bibitem[{Cortes et~al.(2004)Cortes, Haffner, and Mohri}]{cortes_rational_2004}
Corinna Cortes, Patrick Haffner, and Mehryar Mohri. 2004.
\newblock Rational kernels: Theory and algorithms.
\newblock \emph{Journal of Machine Learning Research}, 5:1035--1062.

\bibitem[{Cotterell et~al.(2015)Cotterell, Peng, and
  Eisner}]{cotterell_modeling_2015}
Ryan Cotterell, Nanyun Peng, and Jason Eisner. 2015.
\newblock Modeling word forms using latent underlying morphs and phonology.
\newblock \emph{TACL}, 3:433--447.

\bibitem[{Davidov et~al.(2010)Davidov, Tsur, and
  Rappoport}]{davidov2010enhanced}
Dmitry Davidov, Oren Tsur, and Ari Rappoport. 2010.
\newblock Enhanced sentiment learning using twitter hashtags and smileys.
\newblock In \emph{Proc. of COLING}.

\bibitem[{Dreyer(2011)}]{dreyer_non-parametric_2011}
Markus Dreyer. 2011.
\newblock \emph{A Non-parametric Model for the Discovery of Inflectional
  Paradigms from Plain Text Using Graphical Models over Strings}.
\newblock Ph.D. thesis, Johns Hopkins University.

\bibitem[{Dyer et~al.(2016)Dyer, Kuncoro, Ballesteros, and
  Smith}]{dyer2016rnng}
Chris Dyer, Adhiguna Kuncoro, Miguel Ballesteros, and Noah~A. Smith. 2016.
\newblock Recurrent neural network grammars.
\newblock In \emph{Proc. of NAACL}.

\bibitem[{Eisner(2002)}]{eisner_parameter_2002}
Jason Eisner. 2002.
\newblock Parameter estimation for probabilistic finite-state transducers.
\newblock In \emph{Proc. of ACL}.

\bibitem[{Elman(1990)}]{Elman:1990}
Jeffrey~L. Elman. 1990.
\newblock Finding structure in time.
\newblock \emph{Cognitive science}, 14(2):179--211.

\bibitem[{Foerster et~al.(2017)Foerster, Gilmer, Chorowski, Sohl{-}Dickstein,
  and Sussillo}]{Foerster:2017}
Jakob~N. Foerster, Justin Gilmer, Jan Chorowski, Jascha Sohl{-}Dickstein, and
  David Sussillo. 2017.
\newblock Intelligible language modeling with input switched affine networks.
\newblock In \emph{Proc. of ICML}.

\bibitem[{Forsyth(2016)}]{forsyth2016automatic}
Jonathan~P. Forsyth. 2016.
\newblock \emph{Automatic musical accompaniment using finite state machines}.
\newblock Ph.D. thesis, New York University.

\bibitem[{Gehring et~al.(2017)Gehring, Auli, Grangier, Yarats, and
  Dauphin}]{Gehring:2017}
Jonas Gehring, Michael Auli, David Grangier, Denis Yarats, and Yann Dauphin.
  2017.
\newblock Convolutional sequence to sequence learning.
\newblock In \emph{Proc. of ICML}.

\bibitem[{Ghazvininejad et~al.(2016)Ghazvininejad, Shi, Choi, and
  Knight}]{ghazvininejad2016generating}
Marjan Ghazvininejad, Xing Shi, Yejin Choi, and Kevin Knight. 2016.
\newblock Generating topical poetry.
\newblock In \emph{Proc. of EMNLP}.

\bibitem[{Ghazvininejad et~al.(2017)Ghazvininejad, Shi, Priyadarshi, and
  Knight}]{ghazvininejad2017hafez}
Marjan Ghazvininejad, Xing Shi, Jay Priyadarshi, and Kevin Knight. 2017.
\newblock Hafez: an interactive poetry generation system.
\newblock In \emph{Proc. of ACL, System Demonstrations}.

\bibitem[{Giles et~al.(1992)Giles, Miller, Chen, Chen, Sun, and
  Lee}]{Giles:1992}
C.~Lee Giles, Clifford~B Miller, Dong Chen, Hsing-Hen Chen, Guo-Zheng Sun, and
  Yee-Chun Lee. 1992.
\newblock Learning and extracting finite state automata with second-order
  recurrent neural networks.
\newblock \emph{Neural Computation}, 4(3):393--405.

\bibitem[{Goldberg(2016)}]{Goldberg:2016}
Yoav Goldberg. 2016.
\newblock A primer on neural network models for natural language processing.
\newblock \emph{JAIR}, 57:345--420.

\bibitem[{Hochreiter and Schmidhuber(1997)}]{Hochreiter:1997}
Sepp Hochreiter and J{\"u}rgen Schmidhuber. 1997.
\newblock Long short-term memory.
\newblock \emph{Neural computation}, 9(8):1735--1780.

\bibitem[{Hu and Liu(2004)}]{Hu:2004}
Minqing Hu and Bing Liu. 2004.
\newblock Mining and summarizing customer reviews.
\newblock In \emph{Proc. of KDD}.

\bibitem[{Jordan(1989)}]{jordan1989serial}
Michael~I. Jordan. 1989.
\newblock Serial order: {A} parallel, distributed processing approach.
\newblock In \emph{Advances in Connectionist Theory: {S}peech}. Erlbaum.

\bibitem[{Kingma and Ba(2015)}]{Kingma:2014}
Diederik Kingma and Jimmy Ba. 2015.
\newblock Adam: A method for stochastic optimization.
\newblock In \emph{Proc. of ICLR}.

\bibitem[{Kolen(1993)}]{kolen1993fools}
John~F. Kolen. 1993.
\newblock Fool's gold: Extracting finite state machines from recurrent network
  dynamics.
\newblock In \emph{Proc. of NIPS}.

\bibitem[{Kuich and Salomaa(1986)}]{kuich1985semiring}
Werner Kuich and Arto Salomaa, editors. 1986.
\newblock \emph{Semirings, Automata, Languages}.
\newblock Springer-Verlag.

\bibitem[{Kumar and Byrne(2003)}]{kumar2003weigthed}
Shankar Kumar and William Byrne. 2003.
\newblock A weighted finite state transducer implementation of the alignment
  template model for statistical machine translation.
\newblock In \emph{Proc. of NAACL}.

\bibitem[{LeCun(1998)}]{Lecun:1998}
Yann LeCun. 1998.
\newblock Gradient-based {Learning} {Applied} to {Document} {Recognition}.
\newblock In \emph{Proc. of the {IEEE}}.

\bibitem[{Lee et~al.(2017)Lee, Levy, and Zettlemoyer}]{lee2017ran}
Kenton Lee, Omer Levy, and Luke Zettlemoyer. 2017.
\newblock Recurrent additive networks.
\newblock \emph{arXiv:1705.07393}.

\bibitem[{Lei et~al.(2015)Lei, Barzilay, and Jaakkola}]{Lei:2015}
Tao Lei, Regina Barzilay, and Tommi Jaakkola. 2015.
\newblock {Molding CNNs for text: non-linear, non-consecutive convolutions}.
\newblock In \emph{Proc. of EMNLP}.

\bibitem[{Lei et~al.(2017{\natexlab{a}})Lei, Jin, Barzilay, and
  Jaakkola}]{Lei:2017a}
Tao Lei, Wengong Jin, Regina Barzilay, and Tommi Jaakkola. 2017{\natexlab{a}}.
\newblock Deriving neural architectures from sequence and graph kernels.
\newblock In \emph{Proc. of ICML}.

\bibitem[{Lei et~al.(2016)Lei, Joshi, Barzilay, Jaakkola, Tymoshenko,
  Moschitti, and M\`{a}rquez}]{Lei:2016}
Tao Lei, Hrishikesh Joshi, Regina Barzilay, Tommi Jaakkola, Kateryna
  Tymoshenko, Alessandro Moschitti, and Llu\'{i}s M\`{a}rquez. 2016.
\newblock Semi-supervised question retrieval with gated convolutions.
\newblock In \emph{Proc. of NAACL}.

\bibitem[{Lei et~al.(2017{\natexlab{b}})Lei, Zhang, and Artzi}]{Lei:2017b}
Tao Lei, Yu~Zhang, and Yoav Artzi. 2017{\natexlab{b}}.
\newblock Training {RNNs} as fast as {CNNs}.
\newblock {arXiv}:1709.02755.

\bibitem[{Lei et~al.(2018)Lei, Zhang, Wang, Dai, and Artzi}]{Lei:2018c}
Tao Lei, Yu~Zhang, Sida~I. Wang, Hui Dai, and Yoav Artzi. 2018.
\newblock Simple recurrent units for highly parallelizable recurrence.
\newblock In \emph{Proc. of EMNLP}.

\bibitem[{Leshno and Schocken(1993)}]{leshno1993multilayer}
Moshe Leshno and Shimon Schocken. 1993.
\newblock Multilayer feedforward networks with a nonpolynomial activation
  function can approximate any function.
\newblock \emph{Neural Networks}, 6:861--867.

\bibitem[{Li et~al.(2018)Li, Li, Cook, Zhu, and Gao}]{li2018indrnn}
Shuai Li, Wanqing Li, Chris Cook, Ce~Zhu, and Yanbo Gao. 2018.
\newblock Independently recurrent neural network ({IndRNN}): Building {A}
  longer and deeper {RNN}.
\newblock In \emph{Proc. of CVPR}.

\bibitem[{Lin et~al.(2017)Lin, Li, He, Zhang, and Sun}]{lin2017adversarial}
Kevin Lin, Dianqi Li, Xiaodong He, Zhengyou Zhang, and Ming-Ting Sun. 2017.
\newblock Adversarial ranking for language generation.
\newblock In \emph{Proc. of NIPS}.

\bibitem[{Marcus et~al.(1993)Marcus, Marcinkiewicz, and
  Santorini}]{Marcus:1993}
Mitchell~P. Marcus, Mary~Ann Marcinkiewicz, and Beatrice Santorini. 1993.
\newblock Building a large annotated corpus of {E}nglish: {T}he {P}enn
  {T}reebank.
\newblock \emph{Computational Linguistics}, 19(2):313--330.

\bibitem[{McAuley and Leskovec(2013)}]{McAuley:2013}
Julian McAuley and Jure Leskovec. 2013.
\newblock Hidden factors and hidden topics: understanding rating dimensions
  with review text.
\newblock In \emph{Proc. of RecSys}.

\bibitem[{Melis et~al.(2018)Melis, Dyer, and Blunsom}]{melis2017sota}
G{\'{a}}bor Melis, Chris Dyer, and Phil Blunsom. 2018.
\newblock On the state of the art of evaluation in neural language models.
\newblock In \emph{Proc. of ICLR}.

\bibitem[{Miao et~al.(2016)Miao, Li, Wang, Zhang, and
  Gong}]{miao2016simplifying}
Yajie Miao, Jinyu Li, Yongqiang Wang, Shi-Xiong Zhang, and Yifan Gong. 2016.
\newblock Simplifying long short-term memory acoustic models for fast training
  and decoding.
\newblock In \emph{Proc. of ICASSP}.

\bibitem[{Mikolov et~al.(2014)Mikolov, Joulin, Chopra, Mathieu, and
  Ranzato}]{mikolov2014learning}
Tomas Mikolov, Armand Joulin, Sumit Chopra, Micha{\"{e}}l Mathieu, and
  Marc'Aurelio Ranzato. 2014.
\newblock Learning longer memory in recurrent neural networks.
\newblock \emph{arXiv:1412.7753}.

\bibitem[{Mikolov et~al.(2010)Mikolov, Karafiát, Burget, Cernocký, and
  Khudanpur}]{mikolov2010rnn}
Tomas Mikolov, Martin Karafiát, Lukás Burget, Jan Cernocký, and Sanjeev
  Khudanpur. 2010.
\newblock Recurrent neural network based language model.
\newblock In \emph{Proc. of INTERSPEECH}.

\bibitem[{Mohri et~al.(2002)Mohri, Pereira, and Riley}]{mohri2002weighted}
Mehryar Mohri, Fernando Pereira, and Michael Riley. 2002.
\newblock Weighted finite-state transducers in speech recognition.
\newblock \emph{Computer Speech \& Language}, 16(1):69--88.

\bibitem[{Moore et~al.(2006)Moore, Dines, Magimai-Doss, Vepa, Cheng, and
  Hain}]{moore_juicer_2006}
Darren Moore, John Dines, Mathew Magimai-Doss, Jithendra Vepa, Octavian Cheng,
  and Thomas Hain. 2006.
\newblock Juicer: A weighted finite-state transducer speech decoder.
\newblock In \emph{Proc. of MLMI}.

\bibitem[{Pang and Lee(2004)}]{Pang:2004}
Bo~Pang and Lillian Lee. 2004.
\newblock A sentimental education: Sentiment analysis using subjectivity
  summarization based on minimum cuts.
\newblock In \emph{Proc. of ACL}.

\bibitem[{Pardo and Birmingham(2005)}]{pardo2005modeling}
Bryan Pardo and William Birmingham. 2005.
\newblock Modeling form for on-line following of musical performances.
\newblock In \emph{Proc. of AAAI}.

\bibitem[{Peng et~al.(2017)Peng, Thomson, and Smith}]{peng2017deep}
Hao Peng, Sam Thomson, and Noah~A. Smith. 2017.
\newblock Deep multitask learning for semantic dependency parsing.
\newblock In \emph{Proc. of ACL}.

\bibitem[{Peng et~al.(2018)Peng, Thomson, and Smith}]{peng2018backprop}
Hao Peng, Sam Thomson, and Noah~A. Smith. 2018.
\newblock Backpropagating through structured argmax using a spigot.
\newblock In \emph{Proc. of ACL}.

\bibitem[{Pennington et~al.(2014)Pennington, Socher, and
  Manning}]{Pennington:2014}
Jeffrey Pennington, Richard Socher, and Christopher Manning. 2014.
\newblock Glo{V}e: Global vectors for word representation.
\newblock In \emph{Proc. of EMNLP}.

\bibitem[{Press and Wolf(2017)}]{press2017using}
Ofir Press and Lior Wolf. 2017.
\newblock Using the output embedding to improve language models.
\newblock In \emph{Proc. of EACL}.

\bibitem[{Rastogi et~al.(2016)Rastogi, Cotterell, and
  Eisner}]{rastogi_weighting_2016}
Pushpendre Rastogi, Ryan Cotterell, and Jason Eisner. 2016.
\newblock Weighting finite-state transductions with neural context.
\newblock In \emph{Proc. of NAACL}.

\bibitem[{Sakarovitch(2009)}]{sakarovitch_rational_2009}
Jacques Sakarovitch. 2009.
\newblock Rational and recognisable power series.
\newblock In Manfred Droste, Werner Kuich, and Heiko Vogler, editors,
  \emph{Handbook of Weighted Automata}, pages 105--174. Springer Berlin
  Heidelberg, Berlin, Heidelberg.

\bibitem[{Sch{\"{u}}tzenberger(1961)}]{schutzenberger1961fsa}
Marcel~Paul Sch{\"{u}}tzenberger. 1961.
\newblock On the definition of a family of automata.
\newblock \emph{Information and Control}, 4(2-3):245--270.

\bibitem[{Schwartz et~al.(2018)Schwartz, Thomson, and Smith}]{Schwartz:2018}
Roy Schwartz, Sam Thomson, and Noah~A. Smith. 2018.
\newblock {SoPa}: Bridging {CNNs}, {RNNs}, and weighted finite-state machines.
\newblock In \emph{Proc. of ACL}.

\bibitem[{Siegelmann and Sontag(1995)}]{siegelmann1995rnn}
Hava~T. Siegelmann and Eduardo~D. Sontag. 1995.
\newblock On the computational power of neural nets.
\newblock \emph{Journal of Computer and System Sciences}, 50(1):132--150.

\bibitem[{Socher et~al.(2013)Socher, Perelygin, Wu, Chuang, Manning, Ng, and
  Potts}]{Socher:2013}
Richard Socher, Alex Perelygin, Jean Wu, Jason Chuang, Christopher~D. Manning,
  Andrew Ng, and Christopher Potts. 2013.
\newblock Recursive deep models for semantic compositionality over a sentiment
  treebank.
\newblock In \emph{Proc. of EMNLP}.

\bibitem[{Vaswani et~al.(2017)Vaswani, Shazeer, Parmar, Uszkoreit, Jones,
  Gomez, Kaiser, and Polosukhin}]{vaswani2017attention}
Ashish Vaswani, Noam Shazeer, Niki Parmar, Jakob Uszkoreit, Llion Jones,
  Aidan~N Gomez, \L~ukasz Kaiser, and Illia Polosukhin. 2017.
\newblock Attention is all you need.
\newblock In \emph{Proc. of NIPS}.

\bibitem[{Visser et~al.(2001)Visser, Raijmakers, and Molenaar}]{Visser:2001}
Ingmar Visser, Maartje~EJ Raijmakers, and Peter~CM Molenaar. 2001.
\newblock Hidden markov model interpretations of neural networks.
\newblock In \emph{Connectionist Models of Learning, Development and
  Evolution}, pages 197--206. Springer.

\bibitem[{Weiss et~al.(2018)Weiss, Goldberg, and Yahav}]{weiss2018practical}
Gail Weiss, Yoav Goldberg, and Eran Yahav. 2018.
\newblock On the practical computational power of finite precision {RNNs} for
  language recognition.
\newblock In \emph{Proc. of ACL}.

\bibitem[{Williams and Peng(1990)}]{Williams:1990}
Ronald~J. Williams and Jing Peng. 1990.
\newblock An efficient gradient-based algorithm for online training of
  recurrent network trajectories.
\newblock \emph{Neural computation}, 2(4):490--501.

\bibitem[{Zaremba et~al.(2014)Zaremba, Sutskever, and
  Vinyals}]{zaremba2014recurrent}
Wojciech Zaremba, Ilya Sutskever, and Oriol Vinyals. 2014.
\newblock Recurrent neural network regularization.
\newblock \emph{arXiv:1409.2329}.

\bibitem[{Zoph and Le(2017)}]{zoph2017neural}
Barret Zoph and Quoc~V. Le. 2017.
\newblock Neural architecture search with reinforcement learning.
\newblock In \emph{Proc. of ICLR}.

\end{thebibliography}
\bibliographystyle{acl_natbib_nourl}
\clearpage

\setlength{\abovedisplayskip}{5pt}
\setlength{\belowdisplayskip}{5pt}

\begin{appendices}

\section{Proof of Proposition~\ref{prop:qrnn}}\label{sec:qrnn_proof}
\begin{proof}
	Let's consider a single-layer QRNN with 2-window convolutions:
	\begin{align*}
		\mathbf{f}_t&=\bm{\sigma}\left(\mathbf{V}_f\mathbf{v}_{t-1} + \mathbf{W}_f\mathbf{v}_{t} + \mathbf{b}_f\right),\\
		\mathbf{u}_t&=(\mathbf{1}-\mathbf{f}_t)\odot \bm{g}\bigl(\mathbf{V}_{u}\mathbf{v}_{t-1} + \mathbf{W}_{u} \mathbf{v}_{t} + \mathbf{b}_{u}\bigr),\\
		\mathbf{c}_t &= \mathbf{f}_t\odot \mathbf{c}_{t-1}+\mathbf{u}_{t}.
	\end{align*}
	A similar analysis applies to T-GRUs and T-LSTMs directly,
	and it should be straightforward to generalize the discussion to QRNNs with larger convolution windows.
	
	Let $\Sigma$ denote the alphabet, and let $\mathbf{x} = x_1x_2\dots x_n\in\Sigma^+$ be a nonempty input string.
	consider a WFSA over the real semiring with $2\,\abs{\Sigma} + 1$ states,
	where $q_0$ is the initial state with $\lambda(q_0)=1$;
	$\abs{\Sigma}$ of them are final states $\mathcal{Q}_2=\{q_\alpha\}_{\alpha\in\Sigma}$,
	with $\rho(q_\alpha) = 1$,
	and the remaining $\abs{\Sigma}$ states are denoted by $\mathcal{Q}_1=\{p_\alpha\}_{\alpha\in\Sigma}$.
	
	The transition weights $\tau$ are constructed by
	\sam{a figure would help here.}
	\begin{align*}
		\tau(q_0, p_\alpha, \alpha) &= 1, &&\forall p_\alpha, \in \mathcal{Q}_1;\\
		\tau(p_\alpha, p_\beta, \beta) &= 1, &&\forall p_\alpha, p_\beta \in \mathcal{Q}_1;\\
		\tau(p_\alpha, q_\beta, \beta) &= \mu_\alpha(\beta), &&\forall p_\alpha\in \mathcal{Q}_1, \,\forall  q_\beta\in\mathcal{Q}_2;\\
		\tau(q_\alpha, q_\beta, \beta) &= \phi_\alpha(\beta), &&\forall q_\alpha, q_\beta \in \mathcal{Q}_2.
	\end{align*}
	$\tau=0$ otherwise.
	Then one dimension of the reccurent updates of a 2-window QRNN is recovered by
	parameterizing the weight functions as
	\begin{align}
	\begin{array}{ll}
		\mu_{x_{t-1}}(x_t) = [\mathbf{u}_t]_i, & \phi_{x_{t-1}}(x_t) = [\mathbf{f}_t]_i.
	\end{array}
	\end{align}
	The recurrent computation of a 2-window QRNN of hidden size $d$ can then be recovered by collecting $d$ such WFSAs.
\end{proof}

\section{Proof of Proposition~\ref{proposition:rcnn}}\label{sec:rcnn_proof}
\begin{proof}
	We present the construction of WFSAs for a single layer $n$-gram RCNNs of hidden size $d$.
	
	Let's assume a given input sequence $\mathbf{x}\in\Sigma^+$,
	with $\abs{\mathbf{x}}>n$, since otherwise one only needs include paddings,
	just as in a RCNN.
	Consider a WFSA with $n+1$ states $\mathcal{Q}=\{q_i\}_{i=0}^{n}$
	over the real semiring.
	Use $q_0$ as the initial state with $\lambda(q_0)=1$,
	and $q_n$ as the final state with $\rho(q_n)=1$.
	The transition weight function is defined by 
	\begin{align*}
		\tau(q_i, q_j, \alpha) = 
		\begin{cases}
		1, & i=j=0,\\
		\phi(\alpha), & j=i>0,\\
		\mu_j(\alpha), & j=i+1,\\
		0, & \text{otherwise}.
		\end{cases}
	\end{align*}
	
	Let $z^{(j)}_t$ denote the total score of all paths
	landing in state $q_j$ \textit{just} after consuming $x_t$.
	Let $z^{(j)}_t = 0, j=0,\dots,n$.
	By the forward algorithm
	\begin{align*}
		z^{(0)}_t &= 1,\\
		z^{(j)}_t &= z^{(j)}_{t-1}\,\phi(x_t) + z^{(j-1)}_{t-1}\,\mu_j(x_t), \quad j \geq 1.
	\end{align*}
	Applying similar parametrization to that in \secref{ngram},
	$z^{(n)}_t$ recovers one dimension of the recurrence.
	Collecting $d$ such WFSAs 
	we recover the recurrence of a single layer $n$-gram RCNNs, with $\bm{\lambda}_t$ being a constant,
	or depending only on $x_t$.
\end{proof}

\section{Proof of Proposition~\ref{proposition:isan}}\label{sec:isan_proof}
\begin{proof}
Closely following the 2-dimensional case in \secref{isan},
let's discuss a single layer ISAN of hidden size $d$.

Consider a WFSA over the real semiring with $2d$ states.
Let $d$ of them, denoted by $\mathcal{Q}_2=\{q_i\}_{i=d+1}^{2d}$ be the initial states,
with $\lambda(q_i)=1, i=1,\dots,d$.
Denote the other half $\{q_i\}_{i=1}^{d}$ by $\mathcal{Q}_1$.
Define transition weight $\tau$ by:
\begin{align*}
\tau(q_i, q_j, \alpha) = 
\begin{cases}
1, & i=j, \, q_i\in\mathcal{Q}_2,\\
\eta_j(\alpha), & i=j+d,\\
\mu_{j,i}(\alpha), & q_i,q_j\in\mathcal{Q}_1,\\
0, & \text{otherwise}.
\end{cases}
\end{align*}
$\forall\alpha\in\Sigma$.

Using $q_i\in\mathcal{Q}_1$ as the final state with $\rho(q_i)=1$,
and denote the resulting WFSA by $\wfsa{G}_i$.
By Forward algorithm,
$\wfsa{G}_i$ recovers the $i$th dimension of the single layer ISAN
by letting $[\mathbf{W}_{x_t}]_{i,j}=\mu_{i,j}(x_t)$,
and $[\mathbf{b}_{x_t}]_i=\eta_{i}(x_t)$;
the $d$-dimensional recurrent computation is recovered by 
a set of WFSAs $\{\wfsa{G}_i\}_{i=1}^{d}$ constructed similarly.
\end{proof}

\section{Compared Models}\label{sec:model_details}
This section formally describes the models compared in the experiments (\secref{models}).

\paragraph{\unigram.}
\unigram is derived from $\wfsa{B}$~(\secref{neural}).
	\begin{subequations}
		\begin{align}
		\mathbf{f}_t&=\bm{\sigma}\bigl(\mathbf{W}_{f}\mathbf{v}_{t} + \mathbf{b}_{f}\bigr),\\
		\mathbf{u}_t&=(1-\mathbf{f}_t) \odot\mathbf{W}_{u} \mathbf{v}_{t},\\
		\mathbf{c}_t &= \mathbf{f}_t\odot \mathbf{c}_{t-1}+\mathbf{u}_{t},\\
		\mathbf{o}_t &= \bm{\sigma}\bigl(\mathbf{W}_{o}\mathbf{v}_{t} + \mathbf{b}_{o}\bigr),\label{eq:b:out}\\
		\mathbf{h}_t &= \tanh(\mathbf{o}_t\odot\mathbf{c}_t).
		\end{align}%
	\end{subequations}%

\paragraph{\unimp.} 
Also derived from $\wfsa{B}$, but uses the max-plus semiring~(\secref{semiring}). 
\begin{subequations}
	\begin{align}
	\mathbf{f}_t&=\log\bm{\sigma}\bigl(\mathbf{W}_{f}\mathbf{v}_{t} + \mathbf{b}_{f}\bigr),\\
	\mathbf{u}_t&=\mathbf{W}_{u} \mathbf{v}_{t},\\
	\mathbf{c}_t &= \max\{\mathbf{f}_t + \mathbf{c}_{t-1}, \mathbf{u}_{t}\},\\
	\mathbf{o}_t&=\log\bm{\sigma}\bigl(\mathbf{W}_{o}\mathbf{v}_{t} + \mathbf{b}_{o}\bigr),\label{eq:b_mp:out}\\
	\mathbf{h}_t &= \tanh(\mathbf{o}_{t} + \mathbf{c}_{t}).
	\end{align}
\end{subequations}

\paragraph{\bigram.}  \bigram is derived from $\wfsa{C}$~(\secref{ngram}):
\begin{subequations}
	\begin{align}
	\mathbf{f}^{(j)}_t&=\bm{\sigma}\bigl(\mathbf{W}^{(j)}_{f}\mathbf{v}_{t} + \mathbf{b}^{(j)}_{f}\bigr), \quad j=1,2,\\
	\mathbf{u}^{(j)}_t&=(1-\mathbf{f}^{(j)}_t) \odot\mathbf{W}^{(j)}_{u} \mathbf{v}_{t},\quad j=1,2,\\
	\mathbf{c}_{t}^{(1)} &= \mathbf{c}_{t-1}^{(1)} \odot\mathbf{f}^{(1)}_t
	+ \mathbf{u}^{(1)}_t,\\
	\mathbf{c}_{t}^{(2)} &= \mathbf{c}_{t-1}^{(2)} \odot\mathbf{f}^{(2)}_t 
	+ \mathbf{c}_{t-1}^{(1)}\odot\mathbf{u}^{(2)}_t, \\
	\mathbf{o}_t &= \bm{\sigma}\bigl(\mathbf{W}_{o}\mathbf{v}_{t} + \mathbf{b}_{o}\bigr),\label{eq:c:out}\\
	\mathbf{h}_t &= \tanh(\mathbf{o}_t\odot\mathbf{c}_t).
	\end{align}
\end{subequations}

\paragraph{\both.} Derived from $\wfsa{F}$~(\secref{unigram_bigram}).

\begin{subequations}
	\begin{align}
	\mathbf{f}^{(j)}_t&=\bm{\sigma}\bigl(\mathbf{W}^{(j)}_{f}\mathbf{v}_{t} + \mathbf{b}^{(j)}_{f}\bigr), \quad j=1,2,\\
	\mathbf{u}^{(j)}_t&=(1-\mathbf{f}^{(j)}_t) \odot\mathbf{W}^{(j)}_{u} \mathbf{v}_{t},\quad j=1,2,\\
	\mathbf{p}^{(j)}&=\bm{\sigma}(\mathbf{b}_{p}^{(j)}),\quad j=1,2,\\
	 \mathbf{r}&=\bm{\sigma}(\mathbf{b}_{r}),\\
	\mathbf{c}^{(1)}_t &= \mathbf{c}^{(1)}_{t-1}\odot\mathbf{f}_t^{(1)} + \mathbf{u}_t^{(1)}, \\
	\mathbf{c}^{(2)}_t
	&=\mathbf{c}^{(2)}_{t-1}\odot\mathbf{f}^{(2)}_t 
	+ (\mathbf{c}^{(1)}_{t-1}+\mathbf{r})\odot \mathbf{u}_t^{(2)},\\
	\mathbf{c}_t
	&=\mathbf{p}^{(1)}\odot \mathbf{c}^{(1)}_t + \mathbf{p}^{(2)}\odot\mathbf{c}^{(2)}_t\\
	\mathbf{o}_t &= \bm{\sigma}\bigl(\mathbf{W}_{o}\mathbf{v}_{t} + \mathbf{b}_{o}\bigr),\label{eq:f:out}\\
	\mathbf{h}_t &= \tanh(\mathbf{o}_t\odot\mathbf{c}_t).
	\end{align}
\end{subequations}

The output gates (Equations~\ref{eq:b:out}, \ref{eq:b_mp:out}, \ref{eq:c:out}, and \ref{eq:f:out}) are optional.
They are only used in language modeling experiments,
where we empirically find that they improve performance.

\section{Experimental Setup}
\label{sec:Experimental}
\subsection{Implementation Details}
Our implementation is based on~\citet{Lei:2017b}\footnote{\url{https://github.com/taolei87/sru}}
and~\citet{peng2018backprop},\footnote{\url{https://github.com/Noahs-ARK/SPIGOT}}
using PyTorch.\footnote{\url{https://pytorch.org/}}

\subsection{Language Modeling}
For hyperparameters,
we do not deviate much from the language modeling experiments in~\citet{Lei:2017b}.
We change the hidden sizes for all compared models based on the trainable parameter budget,
and adjust the dropout probabilities accordingly 
to keep the number of remaining hidden units is roughly the same in expectation. 
Besides, we observe that \bigram and \both fail to converge when optimized with 
the SGD algorithm using 1.0 initial learning rate.
And thus we use 0.5 for both models.
Other hyperparameters are kept the same as~\citet{Lei:2017b}.

\newcommand{\ra}[1]{\renewcommand{\arraystretch}{#1}}

\begin{table}[tb]
	\centering
	\ra{1.3}
	\begin{tabular}{lr}
		\toprule
		{\bf Type} & {\bf Values}  \\
		\midrule
		
		Hidden size & $[100,300]$  \\  \midrule
		Vertical dropout & $[0.0,0.5]$ \\ 
		Recurrent dropout&  $[0.0,0.5]$ \\ 
		Embedding dropout& $[0.0,0.5]$ \\\midrule
		Learning Rate & $[10^{-2},10^{-4}]$ \\
		$\ell_2$ regularization & $[10^{-5},10^{-7}]$ \\ \midrule
		Gradient Clipping& $[1.0,5.0]$ \\  
		\bottomrule
	\end{tabular}
	\caption{\label{tab:textcat-Hyperparameters} 
		The hyperparameters explored using random search algorithm in the text classification experiments.}
\end{table}

\subsection{Text classification}

We train our models using Adam \cite{Kingma:2014} with a batch size of 16 (for Amazon) or 64 (for the smaller datasets). Initial learning rate and $\ell_2$ regularization are hyperparameters.
We use 300-dimensional GloVe 840B embeddings \cite{Pennington:2014} normalized to unit length and fixed, replacing unknown words with a special UNK token.
Two layer RNNs are used in all cases.
For regularization, we use three types of dropout: a recurrent variational dropout, vertical dropout and a dropout on the embedding layer.

We tune the hyperparameters of our model on the development set by running 20 epochs of random search.
We then take the best development configuration, and train five models with it using different random seeds. We report the average test results.
The hyperparameters values explored are summarized in \tabref{textcat-Hyperparameters}.
We train all models for 500 epochs, stopping early if development accuracy does not improve for 30 epochs.
During training, we halve the learning rate if development accuracy does not improve for 10 epochs.

\end{appendices}

\end{document}